%% file: iclr2026_conference.tex
\documentclass{article} 
\usepackage{iclr2026_conference,times}

\input{math_commands.tex}

\usepackage{hyperref}
\usepackage{url}

\usepackage[most]{tcolorbox}
\definecolor{DefBlue}{RGB}{231,241,255} 

\newtcolorbox{definitionbox}{
  colback=DefBlue,
  colframe=DefBlue, 
  boxrule=0pt,
  arc=3pt,          
  left=10pt,right=10pt,top=8pt,bottom=8pt,
  enhanced,
  breakable
}
\usepackage{placeins}
\usepackage{graphicx}
\usepackage{wrapfig}
\usepackage{caption}
\usepackage{booktabs} 
\usepackage{amsmath}  
\usepackage{enumitem} 
\usepackage{fancyvrb} 
\usepackage{multirow} 
\usepackage{tcolorbox}
\tcbuselibrary{breakable, skins}

\DeclareCaptionType{sample}[Sample][List of Completions]

\title{Limits of Difficulty Scaling: Hard Samples Yield Diminishing Returns in GRPO-Tuned SLMs}
\iclrfinalcopy

\author{Suraj Yadav,
  Siddharth Yadav\thanks{Equal contribution.},
  Parth Goyal\footnotemark[1] \\
  IIIT Delhi \\
  \texttt{\{suraj24098, siddharth23525, parth23371\}@iiitd.ac.in}
}

\begin{document}

\maketitle

\begin{abstract}
Recent alignment work on Large Language Models (LLMs) suggests preference optimization can improve reasoning by shifting probability mass toward better solutions. We test this claim in a resource-constrained setting by applying GRPO with LoRA to SLMs (0.5B–3B) for math reasoning on GSM8K and MATH datasets with difficulty-stratified analyses. As problem difficulty increases, accuracy plateaus, revealing a capacity boundary: GRPO primarily reshapes output preferences without reliably improving hardest-tier solving. Consistent with this, training GRPO only on lower-difficulty problems matches full-dataset accuracy across difficulty tiers while using only $\sim 45\%$ training steps, indicating diminishing returns from harder samples in this regime. We also find a cross-dataset generalization effect: GSM8K-trained GRPO achieves higher accuracy on the numeric subset of MATH than MATH-trained GRPO, exceeding it by $\sim 5\%$ at 1.5B and by $\sim3\%$ at 3B. We show that the best achievable gains depend strongly on the base model’s prior reasoning competence and the dataset’s difficulty profile.
\end{abstract}

\section{Introduction}

\begin{wrapfigure}{r}{0.6\textwidth}
  \centering
  \vspace{-8pt} 
  \includegraphics[width=\linewidth]{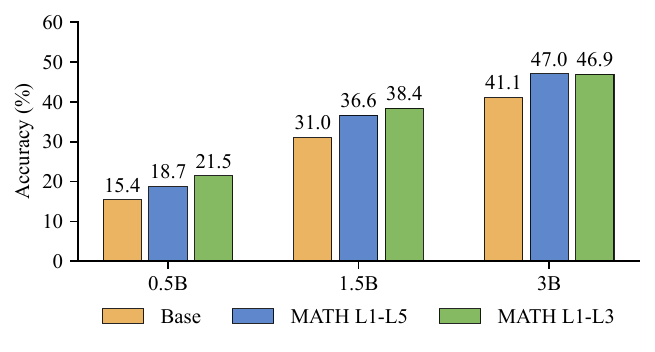}
  \caption{Comparison of accuracy on the MATH dataset for Qwen2.5 0.5B, 1.5B, and 3B models \cite{qwen2025qwen25technicalreport}.
  Models trained on the lower-difficulty subsets (L1–L3) achieve comparable or superior performance to those trained on the full dataset (L1–L5).}
  \label{fig:fig1}
  \vspace{-8pt} 
\end{wrapfigure}

Group Relative Policy Optimization (GRPO) \cite{shao2024deepseekmathpushinglimitsmathematical}, has proven effective for enhancing mathematical reasoning in large language models (LLMs). However, while SLMs often inherit reasoning via distillation from larger teachers \cite{Guo_2025}, the efficacy of applying GRPO directly in resource-constrained regimes remains under-explored. Current literature generally suggests that exposure to higher-difficulty training data is associated with improved reasoning performance \cite{tong2024dartmath}, particularly in LLMs. In compute resource constraint settings, it is critical to determine if Small Language Models (SLMs)  possess the intrinsic 
capacity \cite{wei2022emergentabilitieslargelanguage} 
to benefit from complex alignment samples, or if they face a capacity boundary. 

\begin{definitionbox}
\textbf{Capacity boundary} is a threshold below which algorithmic improvements cannot overcome the fundamental lack of model expressivity required for complex reasoning.
\end{definitionbox}

In this work, we test this hypothesis by conducting a difficulty-stratified \cite{ding2025easy2hardbenchstandardizeddifficultylabels} \cite{gao2025promptcurriculumlearningefficient} analysis of GRPO on SLMs. We fine-tune models ranging from 0.5B to 3B parameters using LoRA \cite{hu2021loralowrankadaptationlarge} on the GSM8K \cite{wei2022emergentabilitieslargelanguage} and MATH \cite{tong2024dartmath} datasets. Unlike standard evaluations, we perform a difficulty-stratified analysis, isolating the impact of training data complexity on downstream performance across different tiers of problem hardness.

\begin{enumerate}
  \item \textbf{Identification of the Capacity Boundary:} We provide empirical evidence that GRPO-based alignment hits a hard saturation point in SLMs, preventing further learning on complex reasoning tasks.

  \item \textbf{Efficiency via Difficulty Stratification:} We demonstrate that filtering out "hard" training samples does not degrade performance. Training on only the lower-difficulty $\sim$ 45\% of the dataset achieves competitive results with full-dataset training, offering a protocol for significantly more efficient alignment.

  \item \textbf{Cross-Dataset Generalization Analysis:} We analyze the transfer capabilities between models trained on GSM8K and MATH, identifying cases where "easier" training distributions (GSM8K) provide a stronger signal for numeric reasoning than domain-specific "hard" distributions (MATH).
  
\end{enumerate}

\section{Related Work}

\subsection{Post-Training Alignment and GRPO}
Post-training alignment methods like RLHF \cite{ouyang2022traininglanguagemodelsfollow}, DPO \cite{rafailov2024directpreferenceoptimizationlanguage}, and specifically Group Relative Policy Optimization (GRPO) \cite{shao2024deepseekmathpushinglimitsmathematical} effectively enhance reasoning in Large Language Models but remain under-explored for Small Language Models (SLMs). Unlike traditional RLHF, which relies on a separate, memory-intensive value function to compute policy baselines, GRPO samples a group of $G$ outputs $\{o_1, \dots, o_G\}$ for a single prompt $q$. It then computes the baseline directly from this group's empirical mean. The optimization objective is formulated as:

\begin{equation}
\begin{split}
\mathcal{J}_{\text{GRPO}}&(\theta) = \mathbb{E}_{q \sim \mathcal{D}, o \sim \pi_{\theta_{\text{old}}}} \\
&\left[ \frac{1}{G} \sum_{i=1}^G \frac{1}{|o_i|} \sum_{t=1}^{|o_i|} \min(r_{i,t}(\theta) \hat{A}_{i,t}, \text{clip}(r_{i,t}(\theta), 1 - \epsilon, 1 + \epsilon)\hat{A}_{i,t}) - \beta D_{\text{KL}}(\pi_\theta \| \pi_{\text{ref}}) \right]
\end{split}
\end{equation}
where 
\[
r_{i,t}(\theta) = \frac{\pi_{\theta}(o_{i,t} \mid q, o_{i,<t})}{\pi_{\theta_{\text{old}}}(o_{i,t} \mid q, o_{i,<t})}.
\]
where the advantage $\hat{A}_{i,t}$ is the normalized reward of output $o_i$ relative to its group. This mechanism significantly reduces memory overhead and naturally incentivizes the model to outperform its own average generation, a dynamic well-suited for iteratively improving mathematical Chain-of-Thought reasoning.

\subsection{Data Complexity and the Capacity Boundary}
Furthermore, while existing curricula prioritize high-complexity data \cite{ding2025easy2hardbenchstandardizeddifficultylabels}, we demonstrate that for capacity-constrained SLMs, such strategies often induce negative transfer. Our work investigates this ``capacity boundary,'' establishing that simpler signals yield superior generalization in sub-3B regimes. (See Appendix~\ref{app:related_work} for extended review).

\section{Methodology}

\paragraph{Training Framework}
We adopt a two-stage training protocol. Models Qwen 2.5 0.5B, 1.5B, 3B \cite{qwen2025qwen25technicalreport} first undergo a brief phase of Supervised Fine-Tuning (SFT) to establish instruction-following behavior and baseline reasoning capabilities. Subsequently, we align these SFT-initialized models using Group Relative Policy Optimization (GRPO) via the TRL library \cite{vonwerra2020trl}. To ensure parameter efficiency, we apply Low-Rank Adaptation (LoRA) to all linear projection layers utilizing the PEFT library \cite{peft}. Detailed hyperparameters for both SFT and GRPO stages are provided in Appendix~\ref{app:repro}.

\paragraph{Difficulty-Aware Reward Modeling}
We design a composite reward function that enforces both strict XML-based formatting and reasoning correctness. A key component of our methodology for the MATH dataset is difficulty scaling: we dynamically weight correctness rewards based on the problem's difficulty level. This ensures that successful reasoning on complex tasks provides a stronger optimization signal.

\paragraph{Dataset \& Stratification}
We utilize the GSM8K dataset and a targeted selection of the Hendrycks MATH benchmark, specifically focusing on four core domains: \textit{algebra}, \textit{counting\_and\_probability}, \textit{geometry}, and \textit{number\_theory}. To verify our hypothesis regarding capacity boundaries, we map these into a unified two-tier complexity hierarchy:
\begin{itemize}[noitemsep, topsep=0pt]
    \item \textbf{Low Difficulty:} GSM8K (grade-school arithmetic), MATH Levels 1--3.
    \item \textbf{High Difficulty:} MATH Levels 4--5.
\end{itemize}

\paragraph{Evaluation Protocol}
We evaluate all models using greedy decoding ($T=0$) to assess deterministic reasoning capabilities. We benchmark performance improvements directly against the \textbf{SFT-tuned baselines} (the model state prior to GRPO alignment). Our critical analysis focuses on \textbf{upward transfer}: measuring whether models trained exclusively on "Low" data partitions can generalize to "High" difficulty test samples without direct exposure to them. Additionally, we extend this evaluation to both GSM8K-trained and MATH-trained models, testing them on the numerical-answer subset of the full MATH test set (see Appendix ~\ref{app:math}).

\section{Experimental Results}

\subsection{The Capacity Boundary: Saturation on High-Difficulty Tasks}
\label{app:4.1}
We investigate whether exposing small models to high-difficulty reasoning paths improves general performance. Figure~\ref{fig:fig1} compares models tuned on the full MATH dataset (Levels 1--5) versus the easier subset (Levels 1--3). We observe that including high-difficulty problems yields \textbf{no significant gain} across any model size. In fact, for the 0.5B and 1.5B models, the inclusion of Level 4--5 data proves detrimental: the 0.5B model achieves \textbf{21.5\% accuracy} on the L1--L3 subset but drops to \textbf{18.7\%} when trained on the full dataset. Similarly, the 1.5B model sees a regression from 38.4\% to 36.6\% with the addition of harder tasks. Even for the 3B model, performance saturates, showing negligible difference between conditions (46.9\% vs. 47.0\%). Crucially, to verify that this saturation is not merely an optimization artifact of parameter-efficient fine-tuning, we conducted a full fine-tuning (FFT) ablation on the 0.5B model. The FFT results mirrored the LoRA findings, showing a similar trend of performance when high-difficulty samples were included (detailed in Appendix~\ref{app:appendix_fft}). This confirms a strict capacity boundary where small models cannot effectively leverage rewards from problems beyond their reasoning horizon. As detailed in Appendix~\ref{app:failure_analysis} Figure~\ref{fig:fig4} and Figure~\ref{fig:fig5}, we observe distinct failure mechanisms dictated by model scale. The 0.5B model simply lacks the capacity to sustain complex inference; it fails to reason deeply, with its average generation length plateauing at approximately \textbf{330 tokens} for Level 5 problems. In contrast, the 1.5B and 3B models actively attempt to solve these tasks, scaling their chain-of-thought to nearly \textbf{400 tokens} for Level 5 problems. However, this extended reasoning leads to \textit{token overflow}: the models frequently "get lost" in their own convoluted reasoning traces or hit length constraints before a final answer can be extracted (See Appendix ~\ref{app:response_analysis} for extended review).

\subsection{Diminishing Returns: Hard Data as Computational Inefficiency}
\label{section:4.2}
\begin{figure}[h] 
    \centering
    \includegraphics[width=\textwidth]{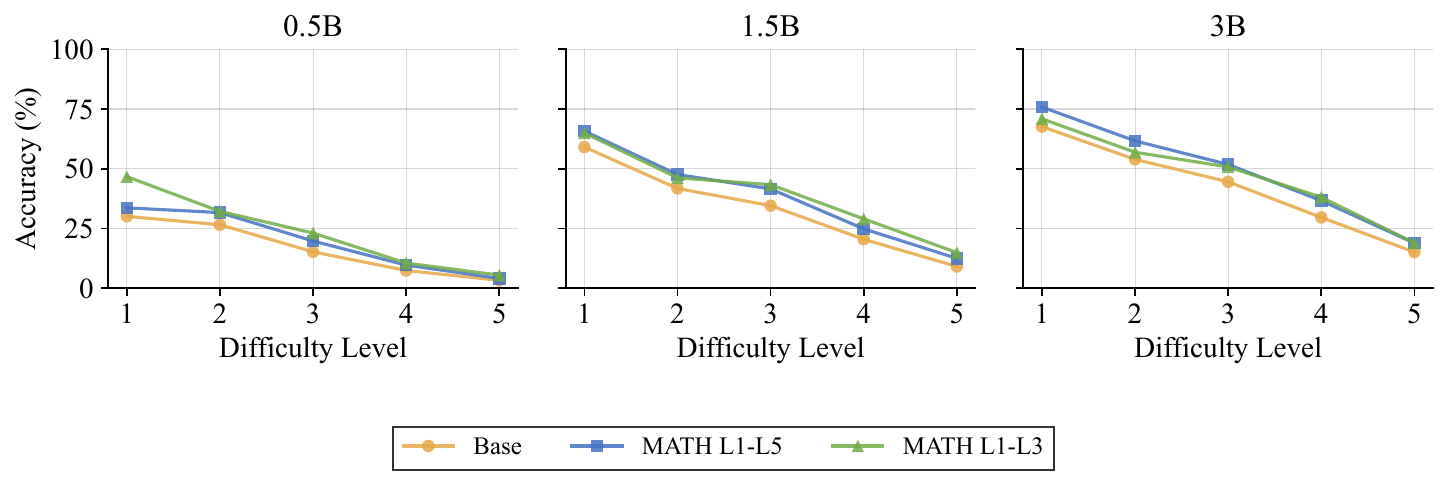} 
    \vspace{-10pt} 
    \caption{Difficulty-stratified MATH accuracy for base and GRPO-tuned models. GRPO trained on full data (L1–L5) performs similarly to GRPO trained on L1–L3, suggesting limited benefit from including higher-difficulty problems (especially for smaller models).}
    \label{fig:fig2}
    \vspace{-10pt} 
\end{figure}

Given the capacity boundary, we investigate the utility of including hard samples in the training distribution. Figure~\ref{fig:fig2} compares models trained on the Full Dataset (MATH L1--L5) versus those trained exclusively on the simple subset (MATH L1--L3).
Remarkably, we observe a signal-to-noise degradation where training on high-difficulty problems degrades performance on those very same problems for smaller models. For the 1.5B model, restricting training to the easy subset (L1–L3) outperforms training on the full dataset (L1–L5) on \textbf{Level 4} tasks (\textbf{29.0\%} vs. \textbf{24.9\%}) and \textbf{Level 5} tasks (\textbf{14.9\%} vs. \textbf{12.5\%}).
To ensure this performance gap is driven by the exclusion of hard samples rather than simply a reduction in dataset size, we trained a control 1.5B model on a random subsample of the full dataset, exactly matched to the size of the L1–L3 split. This size-matched random subset performed even worse than the full dataset, confirming that the presence of high-difficulty samples not the reduced training duration is the root cause of the degradation (detailed in Appendix~\ref{app:appendix_size_matched}).
This suggests that for models below the 3B threshold, unsolvable L5 samples do not provide a learning signal; instead, they introduce high-variance gradients as evidenced by the volatile reward trajectories in Appendix~\ref{app:training_dynamics} Figure~\ref{fig:fig7} that actively interfere with the consolidation of intermediate reasoning patterns.
For the 3B model, the trade-off is distinct: while the Full Dataset yields higher accuracy on Level 1 (75.8\% vs. 70.9\%), the Easy-Only curriculum still achieves superior generalization to Level 4 difficulty (\textbf{38.1\%} vs. \textbf{36.7\%}). Consequently, by filtering these hard samples, we achieve superior alignment outcomes on complex tasks while using only approximately 45\%  of total training steps.

\subsection{Cross-Dataset Generalization: Simpler Signals Transfer Better}

\begin{wrapfigure}{r}{0.6\textwidth}
  \centering
  \vspace{-8pt} 
  \includegraphics[width=\linewidth]{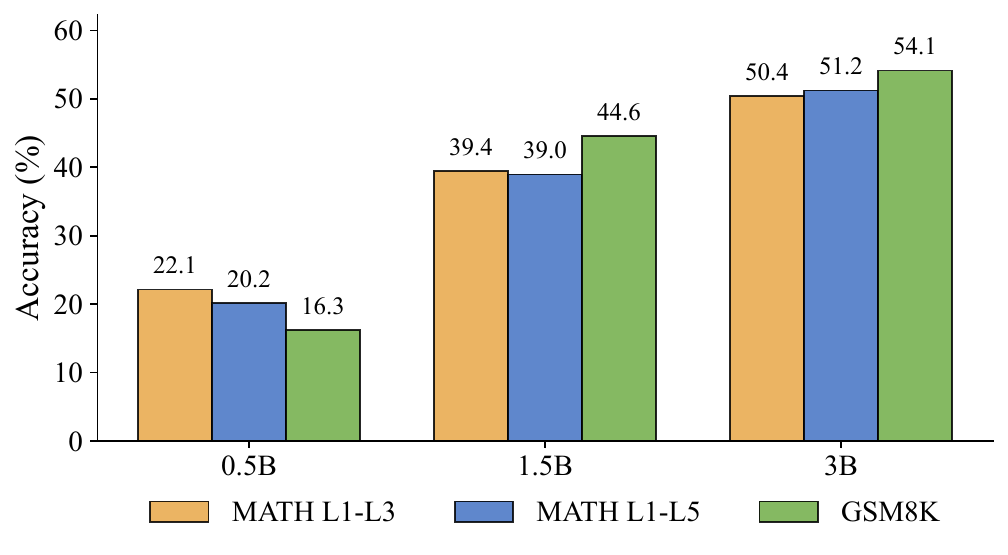}
  \caption{\textbf{Cross-dataset generalization under GRPO.} Accuracy on the \emph{Numerical} subset of MATH for 0.5B/1.5B/3B models trained with GRPO on (i) MATH L1--L3, (ii) MATH L1--L5, and (iii) GSM8K.}
  \label{fig:fig3}
  \vspace{-10pt} 
\end{wrapfigure}

We observe a cross-dataset generalization effect in which training on the simpler GSM8K dataset can transfer better than training on MATH itself. Figure~\ref{fig:fig3} reports the evaluation on the \emph{Numerical} subset of MATH.

The transfer from GSM8K improves with scale: at 1.5B, GSM8K-trained model reaches \textbf{44.6\%} accuracy vs.\ \textbf{39.4\%}/\textbf{39.0\%} for the (MATH L1--L3/L1--L5) models, and at 3B it achieves \textbf{54.1\%} vs.\ \textbf{50.4\%}/\textbf{51.2\%}. Conversely, the 0.5B model does not benefit from this transfer (\textbf{16.3\%} on GSM8K vs.\ \textbf{22.1\%}/\textbf{20.2\%} on MATH), suggesting that very small models need target-dataset supervision (even when both datasets are math-specific).
Appendix~\ref{app:fine_grained} Table~\ref{tab:table2} indicates that these gains come from cleaner transfer in foundational topics. For instance, in \textbf{Algebra}, the GSM8K-trained 1.5B model outperforms the MATH-trained baseline across difficulty (e.g., \textbf{91.2\% vs.\ 81.6\%} at Level~1; \textbf{27.3\% vs.\ 21.7\%} at Level~5), which is consistent with GSM8K providing a lower-variance numeric signal than MATH's heterogeneous objectives.

\section{Discussion \& Limitations}
\paragraph{Interpretation} Our results suggest that for Small Language Models (SLMs), the relationship between training data difficulty and alignment gain is not linear, but rather governed by a signal-to-noise ratio. We hypothesize that "hard" problems require multi-hop reasoning chains that exceed the intrinsic representational capacity of sub-3B models. Consequently, during the exploration phase of GRPO, the model rarely generates a correct reasoning trace for these problems, leading to a sparse reward landscape. Instead of providing a directional gradient toward better reasoning, these hard samples likely contribute high-variance gradients that act as noise, potentially destabilizing the learning of simpler concepts. This explains why removing them and focusing on the "learnable" signal from lower-difficulty tiers maintains performance while using $\sim 45\%$ the training steps.
\paragraph{Limitations} We acknowledge several constraints in our study. First, we employed LoRA (Low-Rank Adaptation) \cite{hu2021loralowrankadaptationlarge} rather than full fine-tuning. While our primary experiments rely on parameter-efficient fine-tuning via LoRA for computational tractability, we verified through a full fine-tuning ablation on the 0.5B model (detailed in Appendix~\ref{app:appendix_fft})) that the capacity boundary persists even with all parameters unfrozen. This confirms the saturation is an intrinsic property of the model's representational scale rather than an optimization constraint. Second, our study is bounded by the scale of standard academic benchmarks (GSM8K \cite{cobbe2021trainingverifierssolvemath}, MATH \cite{hendrycks2021measuringmathematicalproblemsolving}). The dynamics of difficulty scaling might differ in regimes with massive-scale synthetic data \cite{gunasekar2023textbooksneed} (e.g., millions of samples), where brute-force memorization could mimic reasoning. Finally, our findings are specific to the 0.5B -- 3B parameter regime; these trends likely invert for frontier models (70B+), where capacity is sufficient to leverage complex training signals.

\section{Conclusion}
In this work, we challenge the prevailing assumption that "harder is better" for post-training alignment.We demonstrate that for SLMs in resource-constrained settings, data difficulty is not a strict proxy for alignment quality. Our difficulty-stratified analysis reveals a sharp capacity boundary: models in the 0.5B–3B range saturate on complex reasoning tasks, rendering high-difficulty training samples computationally wasteful. Crucially, we show that training on only the lower-difficulty data matches full-dataset performance, offering a clear path to efficient alignment. Our findings suggest a resource-efficient alignment paradigm: train on what the model can learn from, not on what exceeds its capacity.

\bibliography{iclr2026_conference}
\bibliographystyle{iclr2026_conference}

\newpage
\appendix
\section{Extended Related Work}
\label{app:related_work}

Our research sits at the intersection of efficient language modeling, reinforcement learning for mathematical reasoning, and data curriculum design.

\subsection{Alignment for Reasoning}
Standard alignment techniques, such as Reinforcement Learning with Human Feedback (RLHF) \cite{ouyang2022traininglanguagemodelsfollow}, rely on a trained reward model and a value function (critic) to guide generation. While PPO has been fundamental in aligning models like LLaMA-2 \citep{touvron2023llama2openfoundation}, the memory overhead of maintaining a critic model often equal in size to the policy model is prohibitive for efficient training. Direct Preference Optimization (DPO) \citep{rafailov2024directpreferenceoptimizationlanguage} circumvents this by optimizing the policy directly on offline preference pairs. However, for mathematical reasoning, where the solution space is deterministic (binary correctness) but the reasoning path is diverse, online exploration is crucial.

We adopt Group Relative Policy Optimization (GRPO) \citep{shao2024deepseekmathpushinglimitsmathematical}, which specifically targets this domain. GRPO samples a group of outputs for a single prompt and computes the baseline for the policy gradient from the group mean rather than a separate value function. This reduces memory usage and naturally incentivizes the model to outperform its own average generation, a dynamic well-suited for the iterative improvement of Chain-of-Thought reasoning.

\subsection{The Capacity of Small Language Models (SLMs)}
While scaling laws \citep{kaplan2020scalinglawsneurallanguage} have historically driven the development of massive models, recent focus has shifted toward efficiency. The emergence of SLMs, typically under 3B parameters, has demonstrated that high-quality data can offset parameter reduction. The ``Phi'' series \citep{gunasekar2023textbooksneed} validated that models trained on ``textbook-quality'' synthetic data could rival larger models. However, most SLM evaluations focus on Supervised Fine-Tuning (SFT). Our work extends this inquiry into the post-training phase, specifically investigating whether the reasoning capabilities of these compact models can be further elicited through reinforcement learning without inducing collapse.

\subsection{Data Complexity and Curriculum Learning}
The relationship between training data complexity and model performance is a subject of active debate. In mathematical reasoning, recent works argue for ``upward'' complexity scaling. \citet{ding2025easy2hardbenchstandardizeddifficultylabels} and \citet{parashar2025curriculumreinforcementlearningeasy} propose that training on the frontier of difficulty aids in robust generalization. This aligns with findings in larger models where difficult samples provide the gradients necessary to refine complex decision boundaries.

However, our findings suggest a ``Capacity Boundary'' for SLMs. This echoes the theoretical underpinnings of ``grokking'' \citep{power2022grokkinggeneralizationoverfittingsmall}, where certain reasoning behaviors only materialize past a scale threshold. We provide empirical evidence that for sub-3B models, the ``hard'' data widely advocated for in LLM literature functions as high-variance noise, leading to the negative transfer phenomena observed in our experiments.

\section{Detailed Experimental Setup}
\label{app:repro}

We report full training and evaluation details for both datasets used in this work: Hendrycks MATH and GSM8K. Unless explicitly stated, we keep the pipeline and LoRA configuration identical across datasets, varying only dataset-specific prompting, reward computation, and memory-driven batch sizes.

\subsection{Shared Setup Across All Experiments}
\label{app:shared}

\paragraph{Models} We evaluate Qwen2.5 causal language models at three scales: 0.5B, 1.5B, and 3B parameters.

\paragraph{Training Pipeline} All experiments follow a two-stage procedure:
\begin{enumerate}[noitemsep]
    \item \textbf{Supervised Fine-Tuning (SFT):} Using LoRA adapters on the target dataset.
    \item \textbf{GRPO:} Using new LoRA adapters initialized on the SFT-merged checkpoint.
\end{enumerate}

\paragraph{LoRA Configuration} All stages use the same LoRA configuration:
\begin{itemize}[noitemsep, topsep=0pt]
    \item \textbf{Rank/Alpha:} $r=64$, $\alpha=64$.
    \item \textbf{Regularization:} Dropout $0.0$, Bias \texttt{none}.
    \item \textbf{Target Modules:} \texttt{q\_proj, k\_proj, v\_proj, o\_proj, gate\_proj, up\_proj, down\_proj}.
\end{itemize}

\paragraph{Precision and Memory} We conducted all training on a single A100 40GB GPU using BF16 precision. To optimize memory efficiency and throughput, we employed gradient checkpointing (setting \texttt{use\_cache=False}) and utilized xFormers' memory-efficient attention kernels.

\paragraph{GRPO Generations Constraint} GRPO uses $K=4$ sampled rollouts per prompt. We choose the per-device batch size and gradient accumulation steps to ensure the effective batch size is strictly divisible by $K$:
\[
(\texttt{per\_device\_batch} \times \texttt{grad\_accum}) \pmod K \equiv 0.
\]

\subsection{Hendrycks MATH Experiments}
\label{app:math}

\paragraph{Dataset} We use the EleutherAI Hendrycks MATH dataset, aggregating four subsets: \textit{algebra}, \textit{counting\_and\_probability}, \textit{geometry}, and \textit{number\_theory}. We use the official train/test split, allocating 20\% of the training set to SFT and the remaining 80\% to GRPO.
\paragraph{Eval set} We constructed a numerical subset from the full MATH test set by explicitly selecting samples that require integer or fraction answers. Table~\ref{tab:math_stats} provides the detailed sample counts across difficulty levels for both the numerical subset (used for primary evaluation) and the full test set.

\begin{table}[h]
\centering
\small
\caption{Distribution of Hendrycks MATH test samples across difficulty levels.}
\label{tab:math_stats}
\begin{tabular}{l c c c c c c}
\toprule
\textbf{Dataset} & \textbf{Level 1} & \textbf{Level 2} & \textbf{Level 3} & \textbf{Level 4} & \textbf{Level 5} & \textbf{Total} \\
\midrule
\textbf{Numerical} & 214 & 361 & 407 & 466 & 444 & \textbf{1,892} \\
\textbf{Full}      & 242 & 476 & 585 & 661 & 716 & \textbf{2,680} \\
\bottomrule
\end{tabular}
\end{table}

\paragraph{Sequence Lengths} Max prompt length is 768 tokens; max completion length is 768 tokens (SFT uses a 1536 token combined budget).

\paragraph{Hyperparameters}
\begin{itemize}[noitemsep, topsep=0pt]
    \item \textbf{SFT:} LR $2\times 10^{-5}$, 1~epoch, batch size 8, grad accum 4 (cosine schedule, warmup 0.05).
    \item \textbf{GRPO:} LR $3\times 10^{-6}$, 1~epoch, $\beta=0.08$, $\epsilon=0.2$. Weight decay 0.1, max grad norm 0.1.
    \item \textbf{Sampling:} Temp 0.8, top-$p$ 0.95, top-$k$ 0, \texttt{mask\_truncated\_completions=True}.
\end{itemize}

\paragraph{GRPO Batch Scaling} We vary the per-device batch size by model, as shown in Table~\ref{tab:math_batch}.

\begin{table}[h]
\centering
\small
\caption{GRPO batch scaling for Hendrycks MATH. All runs use $K=4$ generations.}
\label{tab:math_batch}
\begin{tabular}{l c c c}
\toprule
\textbf{Model Size} & \textbf{Per-Device Batch} & \textbf{Grad Accum} & \textbf{Effective Batch} \\
\midrule
0.5B & 20 & 1 & 20 \\
1.5B & 16 & 1 & 16 \\
3B   & 12 & 1 & 12 \\
\bottomrule
\end{tabular}
\end{table}

\paragraph{Prompt Template} We use an explicit XML-like structure for answer extraction:

\begin{center}
\fbox{\begin{minipage}{0.9\linewidth}
\small\ttfamily
Solve the math problem.\\
Put reasoning between <reasoning> and </reasoning>.\\
Put ONLY the final answer between <answer> and </answer>.\\
Do not put any other tags inside <answer>...</answer>.\\
\\
Problem:\\
\{question\}\\
\\
Assistant:
\end{minipage}}
\end{center}

\paragraph{Reward Function} The reward is the sum of correctness, formatting, and anti-degeneracy penalties:
\[
\mathcal{R} = \mathcal{R}_{\text{correct}} + \mathcal{R}_{\text{format}} + \mathcal{R}_{\text{trunc}} + \mathcal{R}_{\text{short}}
\]
\begin{description}[leftmargin=*, style=unboxed]
    \item[Correctness ($\mathcal{R}_{\text{correct}}$):] Let $\ell \in \{1,\dots,5\}$ denote the difficulty level. If a valid answer cannot be extracted, we assign a penalty of $-6.0$. Otherwise, we compute equivalence (exact/normalized/SymPy). Rewards are level-weighted:
    \begin{itemize}
        \item \textbf{Correct:} $\{3.0, 3.5, 4.5, 6.0, 8.0\}$ for Levels 1--5.
        \item \textbf{Wrong:} $\{-1.2, -1.4, -1.7, -2.0, -2.3\}$ for Levels 1--5.
    \end{itemize}
    \item[Formatting ($\mathcal{R}_{\text{format}}$):] We add a bonus in range $[-0.20, 0.20]$. The exact structure \texttt{<reasoning>...</reasoning><answer>...</answer>} yields $+0.15$, with minor adjustments ($\pm 0.01$) for tag counts.
    \item[Penalties:] $\mathcal{R}_{\text{trunc}} = -1.0$ if the max token budget is hit without a closed answer. $\mathcal{R}_{\text{short}} = -0.2$ if length $<100$ tokens.
\end{description}

\paragraph{Evaluation} We report greedy decoding accuracy.

\subsection{GSM8K Experiments}
\label{app:gsm8k}

\paragraph{Dataset} We use the main GSM8K dataset, splitting training into 30\% SFT and 70\% GRPO (seed 42).

\paragraph{Sequence Lengths} Max prompt/completion length is 512 tokens for both stages.

\paragraph{Hyperparameters}
\begin{itemize}[noitemsep, topsep=0pt]
    \item \textbf{SFT:} LR $1\times 10^{-5}$, 1~epoch, batch size 8, grad accum 1 (cosine, warmup 0.03).
    \item \textbf{GRPO:} LR $1\times 10^{-5}$, 2~epochs, $\beta=0.02$, $\epsilon=0.2$, \texttt{loss\_type=dr\_grpo}.
    \item \textbf{Sampling:} Temp 1.0, top-$p$ 0.95, top-$k$ 0.
\end{itemize}

\paragraph{GRPO Batch Scaling} Batch sizes are larger relative to MATH, scaled by model size (Table~\ref{tab:gsm8k_batch}).

\begin{table}[h]
\centering
\small
\caption{GRPO batch scaling for GSM8K ($K=4$).}
\label{tab:gsm8k_batch}
\begin{tabular}{l c c c}
\toprule
\textbf{Model Size} & \textbf{Per-Device Batch} & \textbf{Grad Accum} & \textbf{Effective Batch} \\
\midrule
0.5B & 24 & 2 & 48 \\
1.5B & 20 & 2 & 40 \\
3B   & 16 & 2 & 32 \\
\bottomrule
\end{tabular}
\end{table}

\paragraph{Prompt Template} We enforce a single-line final answer:

\begin{center}
\fbox{\begin{minipage}{0.9\linewidth}
\small\ttfamily
Solve the following math word problem.\\
Write a short step-by-step solution .\\
Finish with the final answer on its own line in the exact format:\\
\#\#\#\# <number>\\
\\
Question:\\
\{question\}\\
\\
Answer:
\end{minipage}}
\end{center}

\paragraph{Reward Function} Rewards encourage exact numeric accuracy with continuous relative-error shaping:
\[
\mathcal{R} = \mathcal{R}_{\text{base}} + \mathcal{R}_{\text{format}} + \mathcal{R}_{\text{len}}
\]
\begin{description}[leftmargin=*, style=unboxed]
    \item[Base ($\mathcal{R}_{\text{base}}$):] Let $\hat{y}$ be the prediction and $y$ the gold answer.
    \begin{itemize}
        \item \textbf{Parse Fail:} $-2.0$ (missing delimiter) or $-0.75$ (numeric parse fail).
        \item \textbf{Correct:} $+3.0$ if relative error $< 10^{-6}$.
        \item \textbf{Incorrect:} Continuous penalty scaled by relative error: $-0.5 - 2.5 \cdot \min(1, \text{rel\_err})$.
    \end{itemize}
    \item[Formatting ($\mathcal{R}_{\text{format}}$):] $+0.25$ if the \texttt{\#\#\#\#} delimiter is present.
    \item[Length Penalty ($\mathcal{R}_{\text{len}}$):] Soft penalty for verbosity $>220$ tokens:
    \[
    \mathcal{R}_{\text{len}} = -\min(1.5, \, 0.006 \cdot (\text{tokens} - 220))
    \]
\end{description}

\paragraph{Evaluation} We report greedy accuracy.

\section{Failure Mode Analysis}
\label{app:failure_analysis}

To support the ``Reasoning Collapse'' hypothesis (Section \ref{app:4.1}), we analyze structural correctness and generation behavior across difficulty levels. All results below are from the \textbf{MATH Full}.

\subsection{Structural Collapse: Answer Extraction Failures by Difficulty}
\label{app:failure_format}

\begin{figure}[t]
    \centering
    \includegraphics[width=0.92\linewidth]{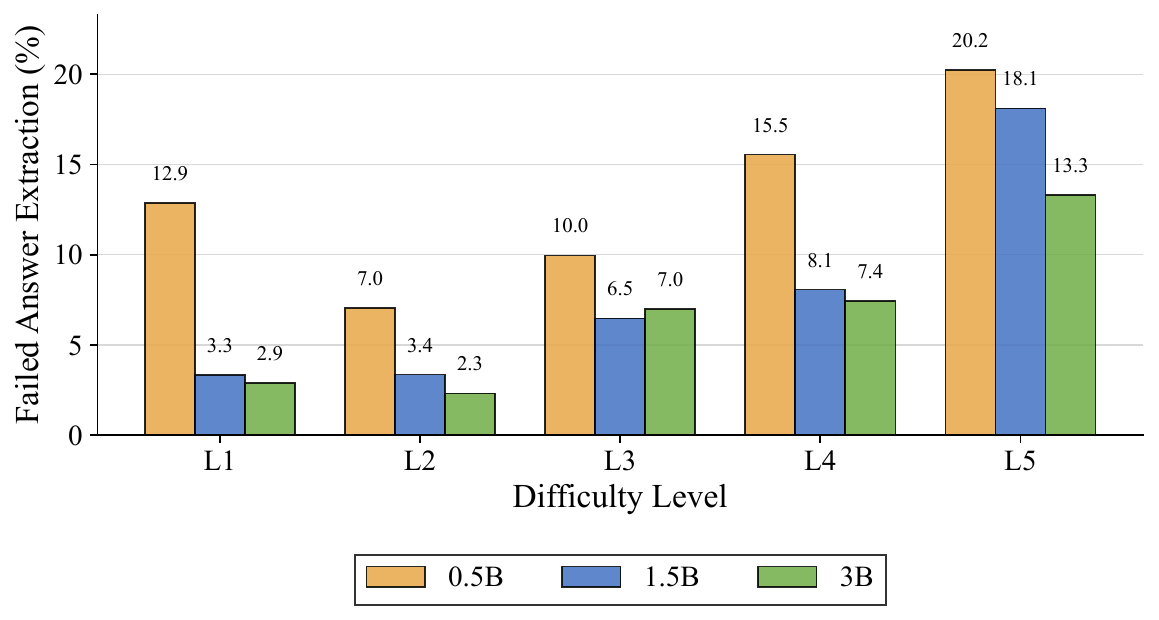}
    \caption{\textbf{Answer Extraction Failure Rate vs. Difficulty (MATH Full).} Percentage of generations without a parsable \texttt{<answer>...</answer>} span. Failures increase with difficulty; the 1.5B model spikes at Level~5 (18.1\%), while 3B remains more stable (13.3\%).}
    \label{fig:fig4}
\end{figure}

We define an \textbf{answer extraction failure} as any generation that does not contain a parsable
\texttt{<answer>...</answer>} span, making the output unscorable by our evaluator.
Figure~\ref{fig:fig4} shows that extraction failures rise with difficulty for all model sizes, with a sharp divergence at Level~5.

Key trends:
\begin{itemize}
    \item \textbf{0.5B:} High failure throughout (e.g., \textbf{12.9\%} at L1 to \textbf{20.2\%} at L5), indicating weak structural control even at easier levels.
    \item \textbf{1.5B:} Relatively controlled at L1--L4 (3.3--8.1\%), but a pronounced \textbf{failure spike at L5} (\textbf{18.1\%}).
    \item \textbf{3B:} More resilient across levels, capping at \textbf{13.3\%} at L5.
\end{itemize}

\subsection{Verifying Reasoning Effort: Average Generation Length}
\label{app:failure_tokens}

\begin{figure}[t]
    \centering
    \includegraphics[width=0.92\linewidth]{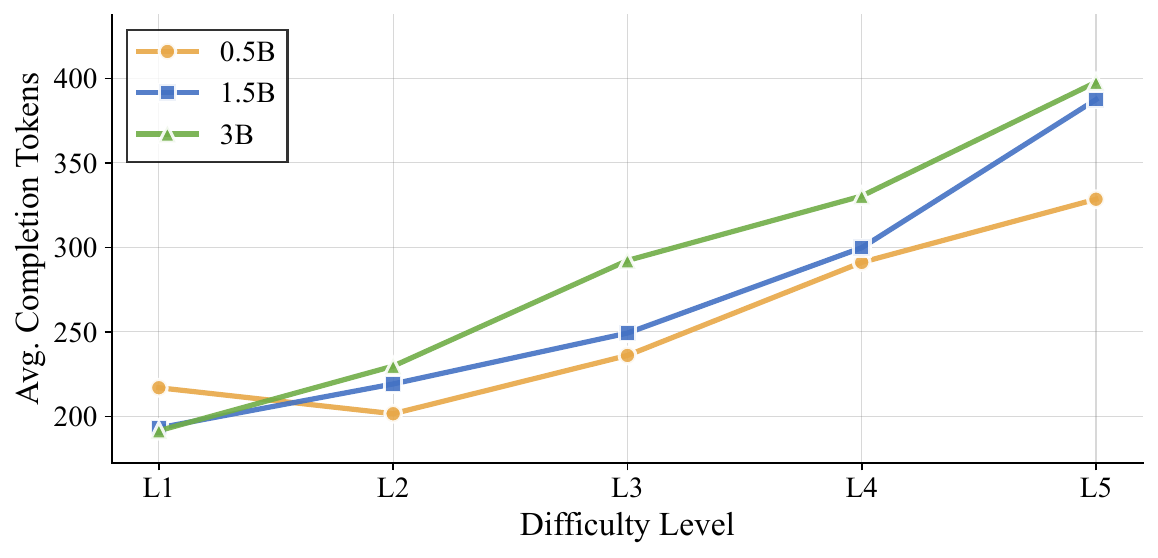}
    \caption{\textbf{Average Completion Tokens vs. Difficulty (MATH Full).} Generation length increases with difficulty across model sizes (approaching $\sim$400 tokens at L5 for 1.5B/3B), indicating sustained reasoning effort even when answer extraction fails.}
    \label{fig:fig5}
\end{figure}

To rule out ``early abandonment,'' we track the average completion length (tokens) per difficulty level.
Figure~\ref{fig:fig5} shows a strong positive relationship between difficulty and generation length for all models.

Taken together, Figures~\ref{fig:fig4} and~\ref{fig:fig5} suggest that failures at high difficulty are not due to shorter outputs, but rather a loss of \emph{end-of-generation structural control} under heavy reasoning load (most evident for the 1.5B model at Level~5).

\newpage
\section{Fine-Grained Performance Analysis}
\label{app:fine_grained}

In this section, we break down model performance by subject and difficulty level to investigate transfer effects across training distributions. We compare models trained on the full dataset (\textbf{MATH}), the Easy subset (\textbf{MATH L1-L3}), and the out-of-domain elementary dataset (\textbf{GSM8K}).

\noindent\textbf{Summary (Table~\ref{tab:table1}).}
Subject-wise accuracy on  (\%) across difficulty Levels L1--L5 for three model sizes (0.5B, 1.5B, 3B), comparing training on \textbf{Full MATH (L1--L5)} versus the \textbf{Easy subset (MATH L1--L3)}.

\noindent\textbf{Summary (Table~\ref{tab:table2}).}
Subject-wise accuracy on numerical subset (\%) across difficulty Levels L1--L5 for three model sizes (0.5B, 1.5B, 3B), comparing three training sources: \textbf{GSM8K}, \textbf{Full MATH}, and \textbf{MATH L1-L3}.

\begin{table*}[t]
\centering
\small
\caption{\textbf{Impact of Filtering Hard Data.} Accuracy (\%) when trained on Full Data (MATH L1--L5) vs.\ Easy Data (MATH L1--L3).}
\label{tab:table1}
\setlength{\tabcolsep}{4pt}
\begin{tabular*}{\textwidth}{@{\extracolsep{\fill}} ll cc c cc c cc}
\toprule
 & & \multicolumn{2}{c}{\textbf{0.5B}} & & \multicolumn{2}{c}{\textbf{1.5B}} & & \multicolumn{2}{c}{\textbf{3B}} \\
\cmidrule{3-4} \cmidrule{6-7} \cmidrule{9-10}
\textbf{Subject} & \textbf{Level} & \textbf{L1-L5} & \textbf{L1-L3} & & \textbf{L1-L5} & \textbf{L1-L3} & & \textbf{L1-L5} & \textbf{L1-L3} \\
\midrule
\multirow{5}{*}{\textbf{Algebra}}
 & L1 & 51.1 & 57.8 & & 83.0 & 78.5 & & 86.7 & 87.4 \\
 & L2 & 38.8 & 41.8 & & 61.7 & 60.2 & & 72.6 & 70.6 \\
 & L3 & 26.8 & 32.6 & & 50.6 & 56.7 & & 67.8 & 66.7 \\
 & L4 & 18.0 & 20.1 & & 44.9 & 45.9 & & 51.6 & 63.6 \\
 & L5 & 10.1 & 12.7 & & 24.8 & 30.6 & & 36.5 & 37.5 \\
\addlinespace
\multirow{5}{*}{\textbf{Count/Prob}}
 & L1 & 30.8 & 43.6 & & 64.1 & 64.1 & & 82.1 & 79.5 \\
 & L2 & 30.7 & 32.7 & & 44.6 & 40.6 & & 64.4 & 55.4 \\
 & L3 & 13.0 & 16.0 & & 42.0 & 42.0 & & 49.0 & 49.0 \\
 & L4 & 6.3  & 6.3  & & 20.7 & 25.2 & & 33.3 & 32.4 \\
 & L5 & 2.4  & 3.3  & & 8.9  & 12.2 & & 14.6 & 12.2 \\
\addlinespace
\multirow{5}{*}{\textbf{Geometry}}
 & L1 & 39.5 & 42.1 & & 52.6 & 47.4 & & 57.9 & 50.0 \\
 & L2 & 35.4 & 29.3 & & 43.9 & 42.7 & & 48.8 & 43.9 \\
 & L3 & 19.6 & 22.5 & & 39.2 & 40.2 & & 46.1 & 43.1 \\
 & L4 & 8.8  & 8.0  & & 12.0 & 23.2 & & 27.2 & 23.2 \\
 & L5 & 0.8  & 1.5  & & 5.3  & 4.5  & & 7.6  & 6.8  \\
\addlinespace
 \multirow{5}{*}{\textbf{Number Theory}}
 & L1 & 13.3 & 43.3 & & 63.3 & 70.0 & & 76.7 & 66.7 \\
 & L2 & 21.7 & 25.0 & & 40.2 & 41.3 & & 60.9 & 57.6 \\
 & L3 & 19.7 & 21.3 & & 34.4 & 34.4 & & 42.2 & 44.3 \\
 & L4 & 5.6 & 7.7 & & 21.8 & 21.8 & & 34.5 & 33.1 \\
 & L5 & 3.2 & 4.5 & & 11.0 & 12.3 & & 16.9 & 19.5 \\

\bottomrule
\end{tabular*}
\end{table*}


\begin{table*}[t]
\centering
\small
\caption{\textbf{Cross-Domain Transfer Analysis.} Accuracy (\%) comparing models trained on GSM8K, MATH (L1-L5), and MATH (L1-L3), evaluated by subject and difficulty.}
\label{tab:table2}
\setlength{\tabcolsep}{3pt}
\begin{tabular*}{\textwidth}{@{\extracolsep{\fill}} ll ccc c ccc c ccc}
\toprule
 & & \multicolumn{3}{c}{\textbf{0.5B}} & & \multicolumn{3}{c}{\textbf{1.5B}} & & \multicolumn{3}{c}{\textbf{3B}} \\
\cmidrule{3-5} \cmidrule{7-9} \cmidrule{11-13}
\textbf{Subject} & \textbf{Level} & \textbf{GSM} & \textbf{L1-L5} & \textbf{L1-L3} & & \textbf{GSM} & \textbf{L1-L5} & \textbf{L1-L3} & & \textbf{GSM} & \textbf{L1-L5} & \textbf{L1-L3} \\
\midrule
\multirow{5}{*}{\textbf{Algebra}}
 & L1 & 61.6 & 53.6 & 56.0 & & 91.2 & 81.6 & 80.8 & & 93.6 & 88.8 & 84.0 \\
 & L2 & 49.1 & 40.9 & 44.0 & & 73.6 & 63.5 & 62.3 & & 83.6 & 76.1 & 72.3 \\
 & L3 & 37.8 & 29.5 & 33.2 & & 67.9 & 51.3 & 57.0 & & 75.6 & 68.9 & 66.8 \\
 & L4 & 21.8 & 19.4 & 18.5 & & 52.6 & 38.9 & 41.2 & & 62.6 & 54.0 & 64.5 \\
 & L5 & 12.1 & 6.1 & 11.6 & & 27.3 & 21.7 & 22.2 & & 34.3 & 29.3 & 31.3 \\
\addlinespace
\multirow{5}{*}{\textbf{Count/Prob}}
 & L1 & 45.2 & 22.6 & 48.4 & & 67.7 & 54.8 & 74.2 & & 74.2 & 77.4 & 77.4 \\
 & L2 & 24.4 & 37.2 & 25.6 & & 44.9 & 50.0 & 38.5 & & 56.4 & 70.5 & 60.3 \\
 & L3 & 15.0 & 10.0 & 10.0 & & 30.0 & 41.7 & 46.7 & & 45.0 & 58.3 & 51.7 \\
 & L4 & 2.9 & 7.1 & 5.7 & & 15.7 & 17.1 & 24.3 & & 34.3 & 28.6 & 24.3 \\
 & L5 & 7.3 & 3.6 & 5.5 & & 9.1 & 9.1 & 5.5 & & 14.5 & 12.7 & 7.3 \\
\addlinespace
\multirow{5}{*}{\textbf{Geometry}}
 & L1 & 44.8 & 48.3 & 34.5 & & 62.1 & 51.7 & 58.6 & & 65.5 & 69.0 & 58.6 \\
 & L2 & 40.0 & 33.3 & 37.8 & & 55.6 & 53.3 & 51.1 & & 62.2 & 62.2 & 60.0 \\
 & L3 & 24.0 & 16.0 & 16.0 & & 46.0 & 52.0 & 46.0 & & 50.0 & 46.0 & 48.0 \\
 & L4 & 6.3 & 6.3 & 6.3 & & 23.8 & 12.7 & 25.4 & & 34.9 & 30.2 & 28.6 \\
 & L5 & 0.0 & 0.0 & 2.0 & & 2.0 & 6.0 & 4.0 & & 12.0 & 10.0 & 10.0 \\
 \addlinespace
 \multirow{5}{*}{\textbf{Number Theory}}
 & L1 & 37.9 & 10.3 & 48.3 & & 72.4 & 69.0 & 58.6 & & 79.3 & 79.3 & 65.5 \\
 & L2 & 32.9 & 20.3 & 22.8 & & 44.3 & 44.3 & 43.0 & & 63.3 & 65.8 & 60.8 \\
 & L3 & 24.0 & 21.2 & 20.2 & & 45.2 & 35.6 & 29.8 & & 57.7 & 50.9 & 47.1 \\
 & L4 & 6.3 & 4.9 & 7.4 & & 22.1 & 23.9 & 19.7 & & 35.2 & 36.0 & 41.0 \\
 & L5 & 1.4 & 2.1 & 2.1 & & 9.9 & 10.6 & 12.1 & & 18.4 & 17.0 & 19.1 \\
\bottomrule
\end{tabular*}
\end{table*}

\section{Training Dynamics \& Stability}
\label{app:training_dynamics}

To verify the optimization stability of Group Relative Policy Optimization (GRPO) across model scales, we analyze the reward trajectories (Mean) and exploration uncertainty (Standard Deviation) throughout the training process.

\subsection{Reward Curves: The Capacity Gap}
\label{app:reward_mean}

\begin{figure}[h]
    \centering
    \includegraphics[width=\textwidth]{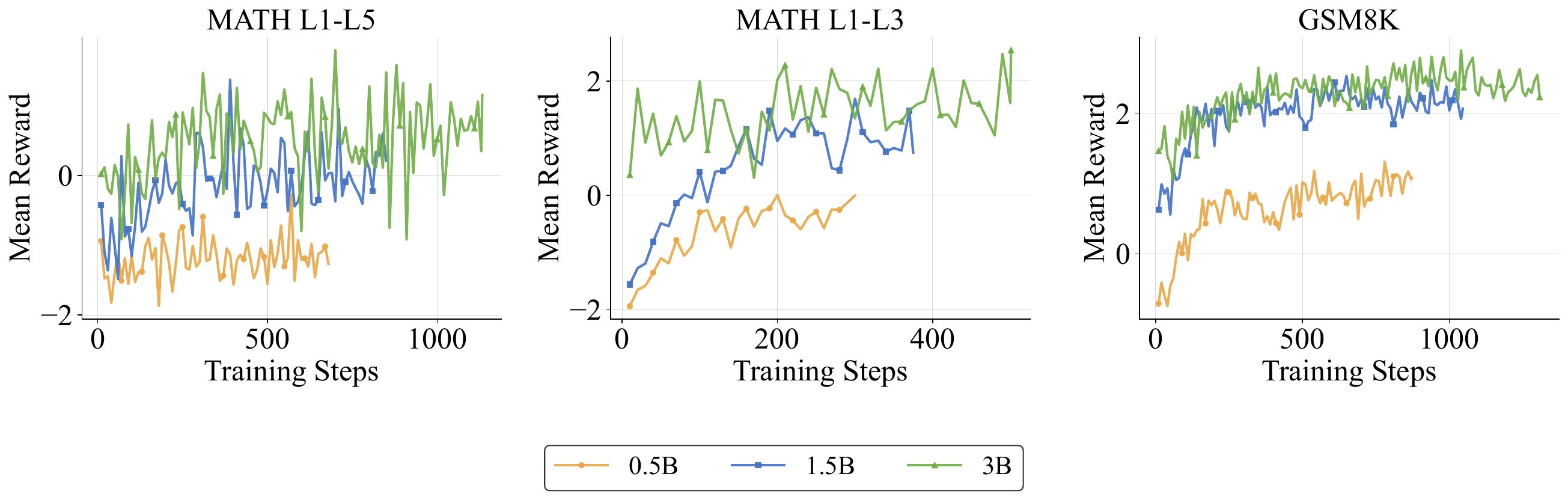} 
    \caption{\textbf{Reward Mean Trajectories.} \textbf{Left/Center:} On the MATH datasets (L1-L5 and L1-L3), the 3B model (Green) demonstrates steady monotonic improvement. In contrast, the 0.5B model (Orange) exhibits a flat or noisy trajectory, indicating an inability to consistently optimize the complex reasoning objective. \textbf{Right:} On the GSM8K dataset, all models—including the 0.5B converge reliably, confirming that the instability is specific to the difficulty of the task relative to model capacity.}
    \label{fig:fig6}
\end{figure}

Figure~\ref{fig:fig6} illustrates the average reward per training step.

The trajectories reveal a distinct split in learning behavior:
\begin{itemize}
    \item \textbf{Convergent Learning (GSM8K):} All models, regardless of size, show a healthy upward trend in reward, validating the training hyperparameters.
    \item \textbf{Capacity Stalling (MATH):} On the harder MATH L1-L5 split, the 0.5B model's reward curve remains near zero or fluctuates randomly. This suggests the model cannot find a gradient direction that consistently leads to correct answers, whereas the 3B model successfully climbs the reward landscape.
\end{itemize}

\subsection{Reward Variance: Guessing vs. Knowing}
\label{app:reward_std}

\begin{figure}[h]
    \centering
    \includegraphics[width=\textwidth]{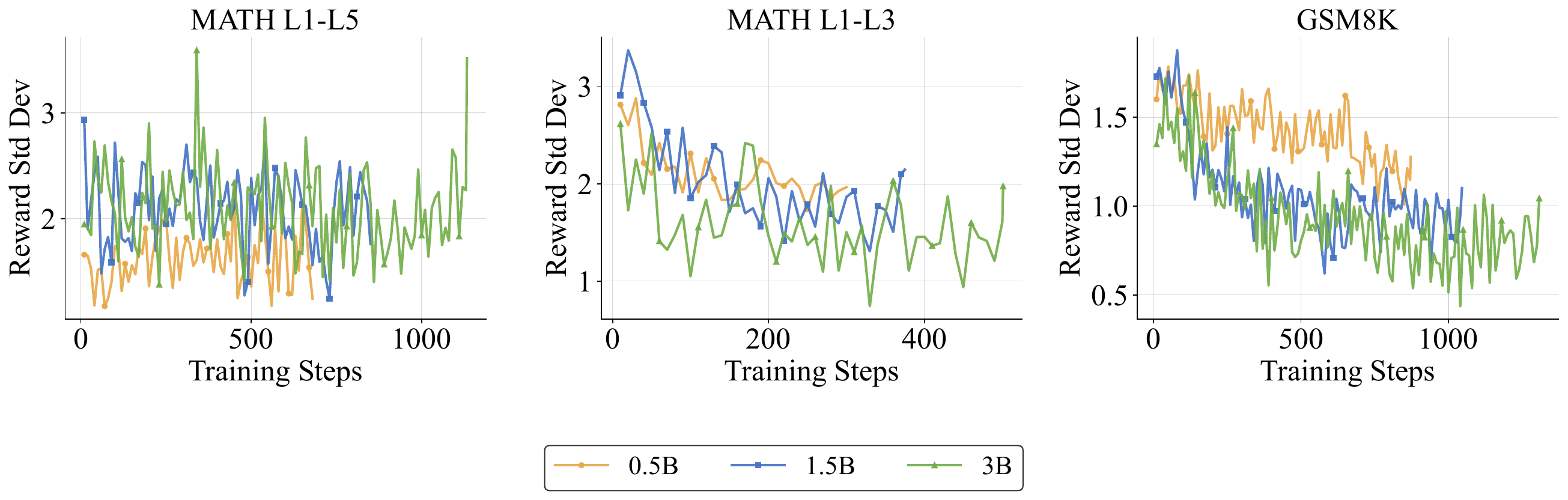} 
    \caption{\textbf{Reward Standard Deviation: } \textbf{Right (GSM8K):} Variance decreases significantly as training progresses (from $\sim 1.8$ to $\sim 0.6$), indicating the models are converging on high-confidence solutions. \textbf{Left (MATH):} Variance remains high and volatile for the 0.5B and 1.5B models, suggesting they are relying on stochastic exploration ("guessing") rather than settling on stable reasoning paths.}
    \label{fig:fig7}
\end{figure}

Figure~\ref{fig:fig7} plots the standard deviation of rewards within each batch, serving as a proxy for model confidence and exploration stability.

\begin{itemize}
    \item \textbf{Healthy Convergence:} On GSM8K, the decreasing standard deviation indicates that the policy is collapsing towards a correct solution mode.
    \item \textbf{High-Variance Exploration:} On MATH L1-L5, the persistent high variance for smaller models implies they are continuously exploring without successfully locking into a high-reward region, effectively "guessing" rather than "learning."
\end{itemize}

\newpage
\section{Full Fine-Tuning (FFT) Ablation}
\label{app:appendix_fft}

\subsection{Motivation}
To ensure that the observed capacity boundary is not merely a consequence of the restricted parameter updates inherent to Low-Rank Adaptation (LoRA), we conducted a full fine-tuning (FFT) ablation on the 0.5B model. This experiment tests whether unfrozen parameters allow Small Language Models (SLMs) to internalize complex logic that parameter-efficient methods might bottleneck.

\begin{wrapfigure}{r}{0.45\textwidth}
\centering
\includegraphics[width=\linewidth]{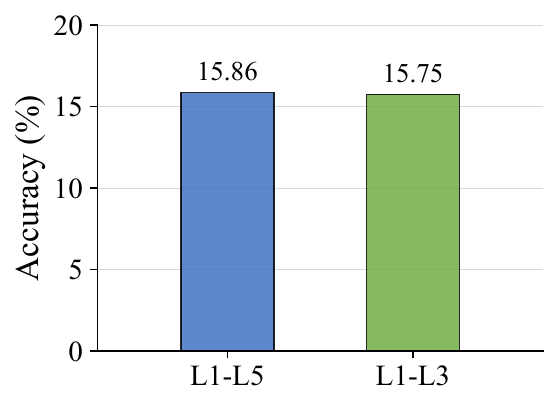} 
\caption{Full Fine-Tuning (FFT) accuracy comparison for 0.5B model on the MATH dataset. Models trained on the easier subset (MATH L1--L3) match the performance of models trained on the full difficulty spectrum (MATH L1--L5), mirroring the LoRA-based capacity boundary.}
\label{fig:fft_comparison}
\end{wrapfigure}

\subsection{Experimental Setup}
We replicated the Group Relative Policy Optimization (GRPO) training pipeline using full parameter updates instead of LoRA adapters. We maintained the exact same hyperparameters as the LoRA-based training, with the exception of the batch sizes. Specifically, batch size was set to 8 for both Supervised Fine-Tuning (SFT) and GRPO stages, and the models were trained for 2 epochs. All other reward formulations, sequence lengths, and difficulty stratification protocols remained identical to the primary MATH setup detailed in Appendix~\ref{app:repro}.

\subsection{Conclusion}
As demonstrated in Figure \ref{fig:fft_comparison}, the fully fine-tuned models trained on the complete dataset (L1--L5) achieve similar performance to their counterparts trained exclusively on the lower-difficulty subset (L1–L3). This confirms that high-difficulty problems actively inject noise into the alignment process for SLMs. Because the performance degradation persists even with all parameters unfrozen, we conclude that the capacity boundary is an intrinsic property of the models' representational scale, not an artifact of the adaptation method.

\section{Control Analysis (Size-Matched Subsampling)}
\label{app:appendix_size_matched}
\subsection{Motivation}
In our primary experiments, the lower-difficulty L1--L3 subset utilizes roughly 45\% of the data compared to the full L1--L5 training set. To ensure the observed performance improvements are not simply a byproduct of training on less data, a control experiment was necessary to completely disentangle the effect of problem difficulty from confounding variables such as dataset size and the total number of optimization steps.

\begin{wrapfigure}{r}{0.45\textwidth}
\centering
\includegraphics[width=\linewidth]{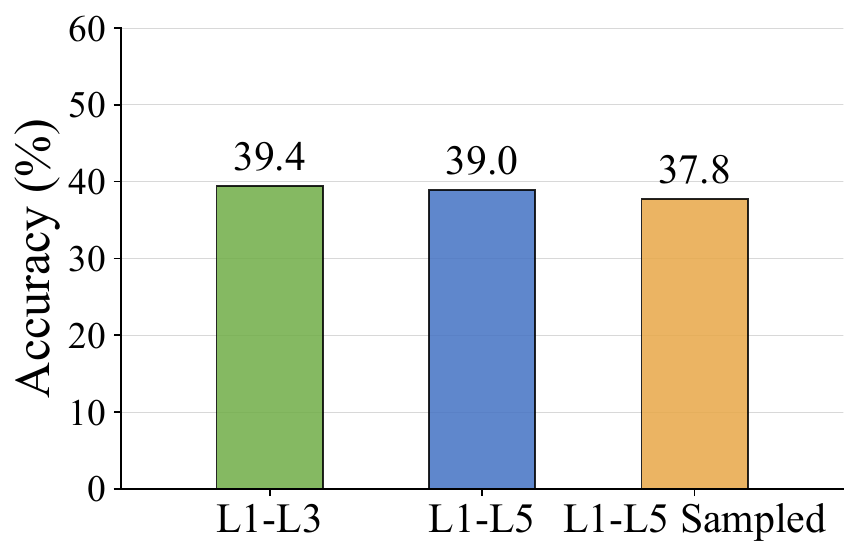} 
\caption{Accuracy comparison of the 1.5B model across three training distributions: the MATH L1--L3 dataset, the MATH L1--L5 subset, and a size-matched random sample of the full dataset MATH L1--L5 (Random 45\%).}
\label{fig:random_subsample}
\end{wrapfigure}

\subsection{Experimental Setup}
To isolate the variable of difficulty, we constructed a size-matched control dataset. We took a random sample of the full MATH L1--L5 dataset, but capped the total sample count to exactly match the size of the L1--L3 subset. We then trained the 1.5B model on this subset using the identical GRPO hyperparameters and LoRA configuration detailed in Appendix \ref{app:repro}.

\subsection{Results and Analysis}

Remarkably, the size-matched Random 45\% subset performed even worse than the full L1--L5 dataset. We hypothesize that this degradation is driven by a deteriorated signal-to-noise ratio. When the model is exposed to the same proportion of ``noisy'' high-difficulty samples, but is given fewer total optimization steps to lock onto the ``clean'' learning signal provided by the easier samples, its ability to consolidate stable reasoning patterns is severely compromised. 

\begin{figure}[h]
\centering
\includegraphics[width=\linewidth]{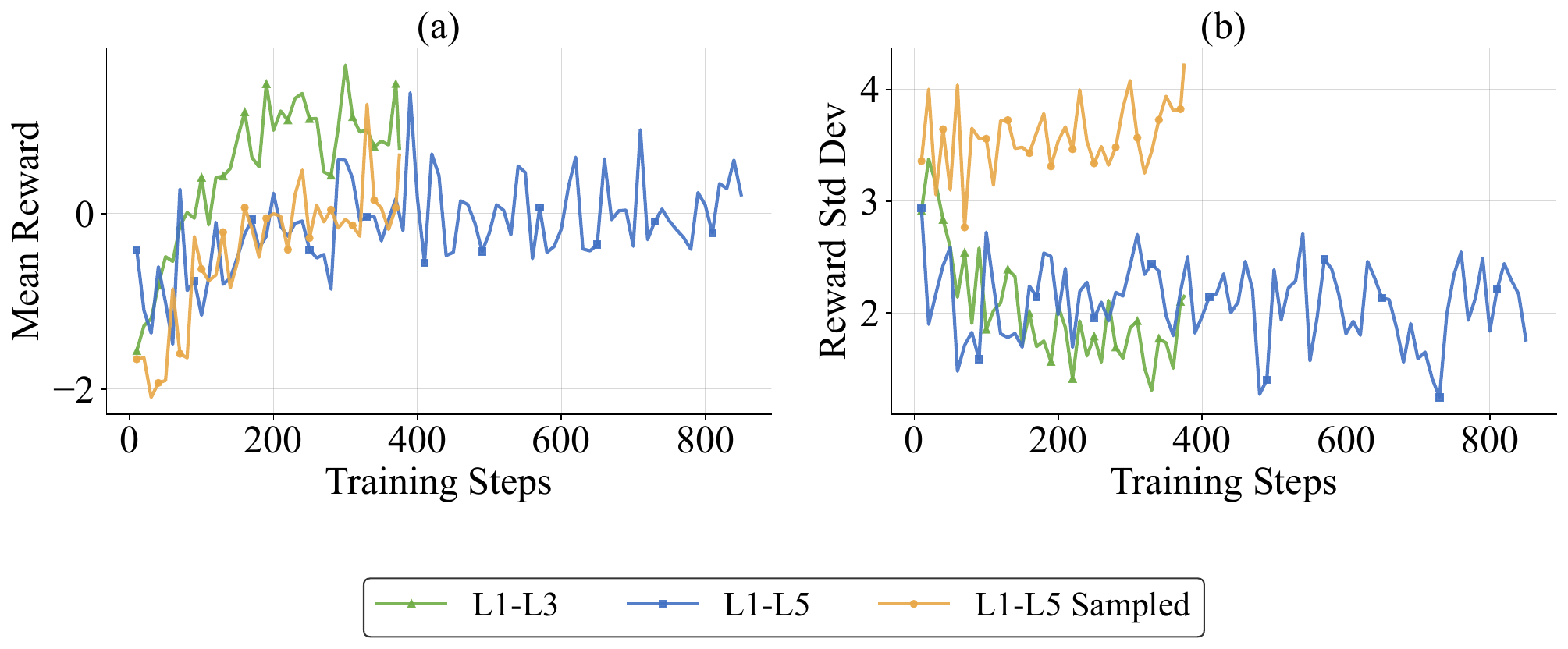}
\caption{Training dynamics for the size-matched Random 45\% control experiment. (a) Mean Reward trajectory. (b) Reward Standard Deviation. The volatile reward and persistently high variance further illustrate the destabilizing effect of hard samples under a reduced step count.}
\label{fig:random_dynamics}
\end{figure}

This instability is explicitly reflected in the training dynamics across the three data splits (Figure \ref{fig:random_dynamics}). While the L1-L3 subset establishes a steady upward gradient, the Mean Reward trajectories for both the full L1-L5 and the Sampled L1-L5 struggle to stabilize, with the Sampled L1-L5 showing the most severe degradation. Concurrently, the Mean Standard Deviation for the Sampled L1-L5 remains persistently higher and more volatile than both the L1-L3 and full L1-L5 runs, indicating that the model is trapped in a state of high-variance exploration without converging on correct reasoning chains.

\subsection{Conclusion}
This confirms that high-difficulty problems actively inject noise into the GRPO optimization process for SLMs, and that our observed gains in Section \ref{section:4.2} are exclusively due to difficulty stratification, not dataset size.

\section{Response Analysis }
\label{app:response_analysis}

\paragraph{Setup} For each training setting, we include two representative generations: one that is correct and one that is incorrect or unscorable. We also provide some observed patterns in the correct and incorrect or unscorable questions.

\paragraph{Observed success patterns.}
Analysis of successful generations reveals a compact set of reusable reasoning \emph{primitives} that consistently yield correct results, provided the task lies within the model's reasoning capabilities:
\begin{itemize}
    \item \textbf{Explicit reduction to canonical forms.}
    Successful generations efficiently map the problem statement into a standard mathematical representation (e.g., factored inequalities, systems of equations, or modulus reductions), allowing for straightforward algorithmic resolution. (Examples: Sample \ref{fig:mathL1-L5_1.5b_success}, Sample \ref{fig:mathL1-L3_0.5b_Success}.)

    \item \textbf{Exhaustive case analysis.}
    When the problem structure requires branching logic (such as absolute values or sign changes), valid solutions systematically enumerate all disjoint cases to ensure complete coverage of the solution space before synthesizing the final result. (Example: Sample \ref{fig:gsm8k_0.5b_success}.)

    \item \textbf{Exploitation of domain invariants.}
    High-fidelity reasoning avoids redundant derivation by leveraging stable domain properties (e.g., angle bisector ratios, area proportionality, or periodicity) to bypass unnecessary symbolic expansion. (Examples: Sample \ref{fig:gsm8k_3b_success}, Sample \ref{fig:mathL1-L3_0.5b_Success}.)

    \item \textbf{Structured constraint modeling.}
    In multi-constraint scenarios, correct solutions are characterized by the formal declaration of variables and the translation of constraints into a coherent system of equations, facilitating systematic elimination. (Examples: Sample \ref{fig:mathL1-L5_3b_Success}, Sample \ref{fig:mathL1-L3_3b_Success}.)

    \item \textbf{Definitive termination.}
    The reasoning process strictly adheres to termination protocols, concluding with a singular, unambiguous output that matches the required format. (Examples: Sample \ref{fig:mathL1-L5_0.5b_success}, Sample \ref{fig:mathL1-L5_1.5b_success}.)
\end{itemize}

\paragraph{Observed failure patterns.}
Analysis of incorrect or unscorable generations reveals that failures converge on specific recurring modes, consistent with a “capacity boundary” interpretation:
\begin{itemize}
    \item \textbf{Cascading arithmetic divergence.}
    Minor numerical errors (e.g., in coefficients or discriminants) frequently alter the mathematical regime of the problem (e.g., shifting from real to complex roots). Instead of backtracking, the model often attempts to force a solution through the invalidated path, leading to incoherent results. (Examples: Sample \ref{fig:gsm8k_3b_failure}, Sample \ref{fig:gsm8k_0.5b_failure}, Sample \ref{fig:mathL1-L3_1.5b_failure}.)

    \item \textbf{Inefficient parameterization vs. structural exploitation.}
    Failures often arise when the model attempts to resolve latent parameters ($a,b,c$) explicitly, rather than identifying necessary invariants or differences. This misallocation of reasoning resources typically results in algebraic bloat or circular logic without progress. (Example: Sample \ref{fig:gsm8k_0.5b_failure}.)

    \item \textbf{Geometric conceptual misalignment.}
    In geometric reasoning tasks, errors frequently stem from an incorrect mapping of textual descriptions to geometric entities, such as conflating an altitude with a radius or misidentifying axes of rotation. (Examples: Sample \ref{fig:mathL1-L5_1.5b_failure}, Sample \ref{fig:mathL1-L5_3b_failure}, Sample \ref{fig:mathL1-L3_3b_failure}.)

    \item \textbf{Hallucinated constraints or entities.}
    In combinatorial contexts, the model occasionally introduces spurious sets or counts not grounded in the prompt. This produces arithmetic that is internally consistent but axiomatically flawed. (Example: Sample \ref{fig:mathL1-L5_0.5b_failure}.)

    \item \textbf{Degenerate repetition and non-termination.}
    A prominent failure signature is the verbatim repetition of reasoning steps, typically emerging when the model encounters an uncertainty barrier. This behavior prevents the generation of a valid final answer span and correlates with high extraction-failure rates in higher-difficulty tiers. (Examples: Sample \ref{fig:gsm8k_3b_failure}, Sample \ref{fig:mathL1-L5_3b_failure}, Sample \ref{fig:mathL1-L3_0.5b_failure}, Sample \ref{fig:mathL1-L3_1.5b_failure}, Sample \ref{fig:mathL1-L3_3b_failure}.)

    \item \textbf{Terminal serialization errors.}
    Certain outputs exhibit valid derivations in the reasoning body but fail at the final extraction layer (e.g., sign omission or formatting inconsistencies). These instances indicate brittleness in the termination protocol rather than a fundamental lack of reasoning capacity. (Example: Sample \ref{fig:gsm8k_1.5b_failure}.)
\end{itemize}

\begin{center}
    \captionof{sample}{Response of 0.5B model trained on GSM8K (success)}
    \label{fig:gsm8k_0.5b_success}
\begin{tcolorbox}[
    enhanced, 
    breakable,             
    colback=white,         
    colframe=black,        
    boxrule=0.5pt,         
    arc=0pt, outer arc=0pt,
    left=1em, right=1em, top=1em, bottom=1em 
]
    \ttfamily 
    \small    
    
    \textbf{Subject:} Algebra (Level 4)
    \vspace{0.5em}
    \hrule
    \vspace{0.5em}

    \textbf{Question:}\\
    What is the sum of the squares of all real values of $x$ for which\\
    $|2 - |x|| = 1$?
    
    \vspace{1em} 
    
    \textbf{Answer:}\\
    First, we need to solve the equation $|2 - |x|| = 1$. 
    We can break this into two cases: $2 - |x| = 1$ and $2 - |x| = -1$.
    For the first case, $2 - |x| = 1$, we get $|x| = 1$, so $x = 1$ or $x = -1$.
    For the second case, $2 - |x| = -1$, we get $|x| = 3$, so $x = 3$ or $x = -3$.
    The real values of $x$ that satisfy the equation are $x = 1$, $x = -1$, $x = 3$, and $x = -3$.
    The sum of the squares of these values is\\[0.5em]
    $\mathtt{1^2 + (-1)^2 + 3^2 + (-3)^2 = 1 + 1 + 9 + 9 = \fbox{20}}$.\\
    
    \#\#\#\# 20

    \vspace{0.5em}
    \hrule
    \vspace{0.5em}
    \textbf{Gold:} 20 \hfill \textbf{Predicted:} 20
\end{tcolorbox}
\end{center}

\begin{center}
    \captionof{sample}{Response of 0.5B model trained on GSM8K (Failure)}
    \label{fig:gsm8k_0.5b_failure}
\begin{tcolorbox}[
    enhanced, 
    breakable,             
    colback=white,         
    colframe=black,        
    boxrule=0.5pt,         
    arc=0pt, outer arc=0pt,
    left=1em, right=1em, top=1em, bottom=1em 
]
    \ttfamily 
    \small    
    
    \textbf{Subject:} Algebra (Level 4)
    \vspace{0.5em}
    \hrule
    \vspace{0.5em}

    \textbf{Question:}\\
    If $\mathtt{f(x)=ax^4-bx^2+x+5}$ and $\mathtt{f(-3)=2,}$ then what is the value of $\mathtt{f(3)}$?
    
    \vspace{1em} 
    
    \textbf{Answer:}\\
    First, we need to find the value of $\mathtt{a}$ using the given information that $\mathtt{f(-3)=2}$.
    Substituting $\mathtt{x=-3}$ into the equation $\mathtt{f(x)=ax^4-bx^2+x+5}$, we get:
    $\mathtt{2=a(-3)^4-b(-3)^2+(3)+5}$
    Simplifying, we have:
    $\mathtt{2=a(81)-9(3)+3+5}$
    $\mathtt{2=a(81)-27+3+5}$
    $\mathtt{2=a(81)-20}$
    $\mathtt{2=a(81)-20}$
    $\mathtt{2=a(81)-20}$
    $\mathtt{2=a(81)-20}$
    $\mathtt{2=a(81)-20}$
    $\mathtt{2=a(81)-20}$
    $\mathtt{2=a(81)-20}$
    $\mathtt{2=a(81)-20}$
    $\mathtt{2=a(81)-20}$
    $\mathtt{2=a(81)-20}$
    $\mathtt{2=a(81)-20}$
    $\mathtt{2=a(81)-20}$
    $\mathtt{2=a(81)-20}$
    $\mathtt{2=a(81)-20}$
    $\mathtt{2=a(81)-20}$
    $\mathtt{2=a(81)-20}$
    $\mathtt{2=a(81)-20}$
    $\mathtt{2=a(81)-20}$
    $\mathtt{2=a(81)-20}$
    $\mathtt{2=a(81)-20}$
    $\mathtt{2=a(81)-20}$
    $\mathtt{2=a(81)-20}$
    $\mathtt{2=a(81)-20}$
    $\mathtt{2=a(81)-20}$
    $\mathtt{2=a(81)-20}$
    $\mathtt{2=a(81)-20}$
    $\mathtt{2=a(81)-20}$
    $\mathtt{2=a(81)-20}$
    $\mathtt{2=a(81)-20}$
    $\mathtt{2=a(81)-20}$
    $\mathtt{2=a(81)-20}$
    $\mathtt{2=a(81)-20}$
    $\mathtt{2=a(81)-20}$
    $\mathtt{2=a(81)-20}$
    $\mathtt{2=a(81)-20}$
    $\mathtt{2=a(81)-20}$
    $\mathtt{2=a(81)-20}$
    $\mathtt{2=a(81)-20}$
    $\mathtt{2=a(81)-20}$
    $\mathtt{2=a(81)-20}$
    $\mathtt{2=a(81)-20}$
    $\mathtt{2=a(81)-20}$
    $\mathtt{2=a(81)-20}$
    $\mathtt{2=a(81)-20}$
    $\mathtt{2=a(81)-20}$
    $\mathtt{2=a(81)-20}$
    $\mathtt{2=a(81)-20}$
    $\mathtt{2=a(81)-20}$
    $\mathtt{2=a(81)-20}$
    $\mathtt{2=a(81)-20}$
    $\mathtt{2=a(81)-20}$
    $\mathtt{2=a(81)-20}$
    $\mathtt{2=a(81)-20}$
    $\mathtt{2=a(81)-20}$
    $\mathtt{2=a(81)-20}$
    $\mathtt{2=a(81)-20}$
    $\mathtt{2=a(81)-20}$
    $\mathtt{2=a(81)-20}$
    
    \vspace{0.5em}
    \hrule
    \vspace{0.5em}
    \textbf{Gold:} 8 \hfill \textbf{Predicted:} -
\end{tcolorbox}
\end{center}

\begin{center}
    \captionof{sample}{Response of 1.5B model trained on GSM8K (success)}
    \label{fig:gsm8k_1.5b_success}
\begin{tcolorbox}[
    enhanced, 
    breakable,             
    colback=white,         
    colframe=black,        
    boxrule=0.5pt,         
    arc=0pt, outer arc=0pt,
    left=1em, right=1em, top=1em, bottom=1em 
]
    \ttfamily 
    \small    
    
    \textbf{Subject:} Algebra (Level 4)
    \vspace{0.5em}
    \hrule
    \vspace{0.5em}

    \textbf{Question:}\\
    Find the constant $c$ such that
    $\mathtt{(x^2-4x+3)(x+5) - (x^2+4x-5)(x-c)=0}$
    for all $x.$
    
    \vspace{1em} 
    
    \textbf{Answer:}\\
    First, we expand the two factors on the left side of the equation:\\
    $\mathtt{(x^2-4x+3)(x+5) = x^3 + 5x^2 - 4x^2 - 20x + 3x + 15 = x^3 + x^2 - 17x + 15}$\\
    $\mathtt{(x^2+4x-5)(x-c) = x^3 - cx^2 + 4x^2 - 4cx - 5x + 5c = x^3 + (4-c)x^2 + (-4c-5)x + 5c}$\\
    Now, we subtract the second factor from the first:\\
    $\mathtt{x^3 + x^2 - 17x + 15 - (x^3 + (4-c)x^2 + (-4c-5)x + 5c) = 0}$\\
    $\mathtt{x^3 + x^2 - 17x + 15 - x^3 - (4-c)x^2 - (-4c-5)x - 5c = 0}$\\
    $\mathtt{x^2 - 17x + 15 - (4-c)x^2 + (4c+5)x - 5c = 0}$\\
    $\mathtt{(-1 + c)x^2 + (4c + 5 - 17)x + (15 - 5c) = 0}$\\
    $\mathtt{(-1 + c)x^2 + (4c - 12)x + (15 - 5c) = 0}$\\
    For this equation to be true for all $x$, the coefficients of $x^2$, $x$, and the constant term must be equal to zero.
    Therefore, we have the following system of equations:\\
    $\mathtt{-1 + c = 0}$\\
    $\mathtt{4c - 12 = 0}$\\
    $\mathtt{15 - 5c = 0}$\\
    Solving the first equation, we find $c = 1$.
    Solving the second equation, we find $c = 3$.
    Solving the third equation, we find $c = 3$.
    Therefore, the only value of $c$ that satisfies all three equations is $\mathtt{c = \fbox{3}}$.\\
    \#\#\#\# 3

    \vspace{0.5em}
    \hrule
    \vspace{0.5em}
    \textbf{Gold:} 3 \hfill \textbf{Predicted:} 3
\end{tcolorbox}
\end{center}

\begin{center}
\captionof{sample}{Response of 1.5B model trained on GSM8K (failure)}
    \label{fig:gsm8k_1.5b_failure}
\begin{tcolorbox}[
    enhanced, 
    breakable,             
    colback=white,         
    colframe=black,        
    boxrule=0.5pt,         
    arc=0pt, outer arc=0pt,
    left=1em, right=1em, top=1em, bottom=1em 
]
    \ttfamily 
    \small    
    
    \textbf{Subject:} Algebra (Level 4)
    \vspace{0.5em}
    \hrule
    \vspace{0.5em}

    \textbf{Question:}\\
    What value of $x$ will give the minimum value for $\mathtt{9x^2 + 18x + 7}$?
    
    \vspace{1em} 
    
    \textbf{Answer:}\\
    To find the minimum value of the quadratic function $\mathtt{9x^2 + 18x + 7}$, we can use the formula for the vertex of a parabola, which is given by $\mathtt{x = -\frac{b}{2a}}$, where $a$ and $b$ are the coefficients of the quadratic function.\\
    In this case, $\mathtt{a = 9}$ and $\mathtt{b = 18}$, so we have $\mathtt{x = -\frac{18}{2(9)} = -\frac{18}{18} = -1}$.\\
    Therefore, the value of $x$ that will give the minimum value for $\mathtt{9x^2 + 18x + 7}$ is $\mathtt{\fbox{-1}}$.\\
    \#\#\#\# 1

    \vspace{0.5em}
    \hrule
    \vspace{0.5em}
    \textbf{Gold:} -1 \hfill \textbf{Predicted:} 1
\end{tcolorbox}
\end{center}

\begin{center}
\captionof{sample}{Response of 3B model trained on GSM8K (success)}
    \label{fig:gsm8k_3b_success}
\begin{tcolorbox}[
    enhanced, 
    breakable,             
    colback=white,         
    colframe=black,        
    boxrule=0.5pt,         
    arc=0pt, outer arc=0pt,
    left=1em, right=1em, top=1em, bottom=1em 
]
    \ttfamily 
    \small    
    
    \textbf{Subject:} Geometry (Level 5)
    \vspace{0.5em}
    \hrule
    \vspace{0.5em}

    \textbf{Question:}\\
    Let $\mathtt{\triangle ABC}$ be a triangle such that $\mathtt{AB = 13,}$ $\mathtt{BC = 14,}$ and $\mathtt{CA = 15.}$ Meanwhile, $D$ is a point on $\mathtt{BC}$ such that $\mathtt{AD}$ bisects $\mathtt{\angle A.}$ Find the area of $\mathtt{\triangle ADC.}$
    
    \vspace{1em} 
    
    \textbf{Answer:}\\
    By the Angle Bisector Theorem, we have $\mathtt{\frac{BD}{DC} = \frac{AB}{AC} = \frac{13}{15}.}$\\
    Let $\mathtt{BD = 13x}$ and $\mathtt{DC = 15x.}$ Then $\mathtt{BD + DC = BC = 14,}$ so $\mathtt{13x + 15x = 14,}$ which gives $\mathtt{x = \frac{1}{2}.}$\\
    Therefore, $\mathtt{BD = 13x = \frac{13}{2}}$ and $\mathtt{DC = 15x = \frac{15}{2}.}$\\
    By Heron's formula, the area of $\mathtt{\triangle ABC}$ is $\mathtt{\sqrt{s(s-AB)(s-BC)(s-CA)},}$ where $s$ is the semiperimeter.\\
    The semiperimeter is $\mathtt{s = \frac{AB + BC + CA}{2} = \frac{13 + 14 + 15}{2} = 21.}$\\
    So the area of $\mathtt{\triangle ABC}$ is $\mathtt{\sqrt{21(21-13)(21-14)(21-15)} = \sqrt{21 \cdot 8 \cdot 7 \cdot 6} = 84.}$\\
    Since $\mathtt{AD}$ bisects $\mathtt{\angle A,}$ the area of $\mathtt{\triangle ADC}$ is $\mathtt{\frac{DC}{BC} \cdot \text{Area of } \triangle ABC = \frac{15}{28} \cdot 84 = \fbox{45}.}$\\
    \#\#\#\# 45

    \vspace{0.5em}
    \hrule
    \vspace{0.5em}
    \textbf{Gold:} 45 \hfill \textbf{Predicted:} 45
\end{tcolorbox}
\end{center}

\begin{center}
\captionof{sample}{Response of 3B model trained on GSM8K (Failure)}
    \label{fig:gsm8k_3b_failure}
\begin{tcolorbox}[
    enhanced, 
    breakable,             
    colback=white,         
    colframe=black,        
    boxrule=0.5pt,         
    arc=0pt, outer arc=0pt,
    left=1em, right=1em, top=1em, bottom=1em 
]
    \ttfamily 
    \small    
    
    \textbf{Subject:} Algebra (Level 4)
    \vspace{0.5em}
    \hrule
    \vspace{0.5em}

    \textbf{Question:}\\
    The equation $\mathtt{y = -6t^2 + 43t}$ describes the height (in feet) of a projectile $t$ seconds after it is launched from the surface of Mars at 43 feet per second. In how many seconds will the projectile first reach 77 feet in height? Express your answer as a decimal rounded to the nearest tenth.
    
    \vspace{1em} 
    
    \textbf{Answer:}\\
    We need to find the time $t$ when the height $y$ is 77 feet.\\
    So we set $\mathtt{y = 77}$ and solve for $t$:\\
    $\mathtt{77 = -6t^2 + 43t}$\\
    Rearranging the equation, we get:\\
    $\mathtt{6t^2 - 43t + 77 = 0}$\\
    We can solve this quadratic equation using the quadratic formula:\\
    $\mathtt{t = \frac{-b \pm \sqrt{b^2 - 4ac}}{2a}}$\\
    where $\mathtt{a = 6}$, $\mathtt{b = -43}$, and $\mathtt{c = 77}$.\\
    Plugging in the values, we get:\\
    $\mathtt{t = \frac{43 \pm \sqrt{(-43)^2 - 4(6)(77)}}{2(6)}}$\\
    $\mathtt{t = \frac{43 \pm \sqrt{1849 - 1872}}{12}}$\\
    $\mathtt{t = \frac{43 \pm \sqrt{1849 - 1872}}{12}}$\\
    $\mathtt{t = \frac{43 \pm \sqrt{1849 - 1872}}{12}}$\\
    $\mathtt{t = \frac{43 \pm \sqrt{1849 - 1872}}{12}}$\\
    $\mathtt{t = \frac{43 \pm \sqrt{1849 - 1872}}{12}}$\\
    $\mathtt{t = \frac{43 \pm \sqrt{1849 - 1872}}{12}}$\\
    $\mathtt{t = \frac{43 \pm \sqrt{1849 - 1872}}{12}}$\\
    $\mathtt{t = \frac{43 \pm \sqrt{1849 - 1872}}{12}}$\\
    $\mathtt{t = \frac{43 \pm \sqrt{1849 - 1872}}{12}}$\\
    $\mathtt{t = \frac{43 \pm \sqrt{1849 - 1872}}{12}}$\\
    $\mathtt{t = \frac{43 \pm \sqrt{1849 - 1872}}{12}}$\\
    $\mathtt{t = \frac{43 \pm \sqrt{1849 - 1872}}{12}}$\\
    $\mathtt{t = \frac{43 \pm \sqrt{1849 - 1872}}{12}}$\\
    $\mathtt{t = \frac{43 \pm \sqrt{1849 - 1872}}{12}}$\\
    $\mathtt{t = \frac{43 \pm \sqrt{1849 - 1872}}{12}}$\\
    $\mathtt{t = \frac{43 \pm \sqrt{1849 - 1872}}{12}}$\\
    $\mathtt{t = \frac{43 \pm \sqrt{1849 - 1872}}{12}}$\\
    $\mathtt{t = \frac{43 \pm \sqrt{1849 - 1872}}{12}}$\\
    $\mathtt{t = \frac{43 \pm \sqrt{1849 - 1872}}{12}}$\\
    $\mathtt{t = \frac{43 \pm \sqrt{1849 - 1872}}{12}}$\\
    $\mathtt{t = \frac{43 \pm \sqrt{1849 - 1872}}{12}}$

    \vspace{0.5em}
    \hrule
    \vspace{0.5em}
    \textbf{Gold:} 3.5 \hfill \textbf{Predicted:} -
\end{tcolorbox}
\end{center}

\begin{center}
    \captionof{sample}{Response of 0.5B model trained on MATH L1-L5 (success)}
    \label{fig:mathL1-L5_0.5b_success}
\begin{tcolorbox}[
    enhanced, 
    breakable,             
    colback=white,         
    colframe=black,        
    boxrule=0.5pt,         
    arc=0pt, outer arc=0pt,
    left=1em, right=1em, top=1em, bottom=1em 
]
    \ttfamily 
    \small    
    
    \textbf{Subject:} Algebra (Level 4)
    \vspace{0.5em}
    \hrule
    \vspace{0.5em}

    \textbf{Question:}\\
    If $\mathtt{f(x) = \frac{4x+1}{3}}$ what is the value of $\mathtt{\left[f^{-1}(1)\right]^{-1}}$?
    
    \vspace{1em} 
    
    \textbf{Answer:}\\
    <reasoning>\\
    We want to find the value of $\mathtt{[f^{-1}(1)]^{-1}}$, which means we want to find the value of $\mathtt{x}$ such that $\mathtt{f(x) = 1}$.\\
    We know that $\mathtt{f(x) = \frac{4x+1}{3}}$, so we set $\mathtt{\frac{4x+1}{3} = 1}$ and solve for $\mathtt{x}$.\\
    Multiplying both sides by 3, we get $\mathtt{4x+1 = 3}$.\\
    Subtracting 1 from both sides, we get $\mathtt{4x = 2}$.\\
    Dividing both sides by 4, we get $\mathtt{x = \frac{1}{2}}$.\\
    Therefore, $\mathtt{[f^{-1}(1)]^{-1} = \frac{1}{1/2} = \fbox{2}}$.\\
    </reasoning>\\
    <answer>2</answer>

    \vspace{0.5em}
    \hrule
    \vspace{0.5em}
    \textbf{Gold:} 2 \hfill \textbf{Predicted:} 2
\end{tcolorbox}
\end{center}

\begin{center}
\captionof{sample}{Response of 0.5B model trained on MATH L1-L5 (Failure)}
    \label{fig:mathL1-L5_0.5b_failure}
\begin{tcolorbox}[
    enhanced, 
    breakable,             
    colback=white,         
    colframe=black,        
    boxrule=0.5pt,         
    arc=0pt, outer arc=0pt,
    left=1em, right=1em, top=1em, bottom=1em 
]
    \ttfamily 
    \small    
    
    \textbf{Subject:} Counting and Probability (Level 4)
    \vspace{0.5em}
    \hrule
    \vspace{0.5em}

    \textbf{Question:}\\
    Our physics club has 20 members, among which we have 3 officers: President, Vice President, and Treasurer. However, one member, Alex, hates another member, Bob. How many ways can we fill the offices if Alex refuses to serve as an officer if Bob is also an officer? (No person is allowed to hold more than one office.)
    
    \vspace{1em} 
    
    \textbf{Answer:}\\
    <reasoning>\\
    Let $\mathtt{A}$ be the set of members who hate Bob, and $\mathtt{B}$ be the set of members who hate Alex. Then $\mathtt{|A| = 15}$ and $\mathtt{|B| = 15}$. We want to count the number of ways to assign the members of $\mathtt{A}$ and $\mathtt{B}$ to the three positions, and then subtract the number of ways to assign the members of $\mathtt{A}$ and $\mathtt{B}$ to the three positions, where the first member of $\mathtt{A}$ is assigned to the President's position, the second member of $\mathtt{A}$ is assigned to the Vice President's position, and the third member of $\mathtt{A}$ is assigned to the Treasurer's position. There are 15 ways to assign the members of $\mathtt{A}$ to the three positions, and 15 ways to assign the members of $\mathtt{B}$ to the three positions, so there are $\mathtt{15 \cdot 15 = \fbox{225}}$ ways to assign the members of $\mathtt{A}$ and $\mathtt{B}$ to the three positions. However, we must also subtract the number of ways to assign the members of $\mathtt{A}$ and $\mathtt{B}$ to the three positions, where the first member of $\mathtt{A}$ is assigned to the President's position, the second member of $\mathtt{A}$ is assigned to the Vice President's position, and the third member of $\mathtt{A}$ is assigned to the Treasurer's position. There are 15 ways to assign the members of $\mathtt{A}$ to the three positions, and 15 ways to assign the members of $\mathtt{B}$ to the three positions, so there are $\mathtt{15 \cdot 15 = \fbox{225}}$ ways to assign the members of $\mathtt{A}$ and $\mathtt{B}$ to the three positions. However, we must also subtract the number of ways to assign the members of $\mathtt{A}$ and $\mathtt{B}$ to the three positions, where the first member of $\mathtt{A}$ is assigned to the President's position, the second member of $\mathtt{A}$ is assigned to the Vice President's position, and the third member of $\mathtt{A}$ is assigned to the Treasurer's position. There are 15 ways to assign the members of $\mathtt{A}$ to the three positions, and 15 ways to assign the members of $\mathtt{B}$ to the three positions, so there are $\mathtt{15 \cdot 15 = \fbox{225}}$ ways to assign the members of $\mathtt{A}$ and $\mathtt{B}$ to the three positions.\\
    </reasoning>\\
    <answer>225</answer>

    \vspace{0.5em}
    \hrule
    \vspace{0.5em}
    \textbf{Gold:} 6732 \hfill \textbf{Predicted:} 225
\end{tcolorbox}
\end{center}

\begin{center}
\captionof{sample}{Response of 1.5B model trained on MATH L1-L5 (Success)}
    \label{fig:mathL1-L5_1.5b_success}
\begin{tcolorbox}[
    enhanced, 
    breakable,             
    colback=white,         
    colframe=black,        
    boxrule=0.5pt,         
    arc=0pt, outer arc=0pt,
    left=1em, right=1em, top=1em, bottom=1em 
]
    \ttfamily 
    \small    
    
    \textbf{Subject:} Algebra (Level 4)
    \vspace{0.5em}
    \hrule
    \vspace{0.5em}

    \textbf{Question:}\\
    Let $a$ be the smallest integer satisfying the inequality $x^2 - 15 < 2x$, and let $b$ be the largest integer satisfying the same inequality. What is $b-a$?
    
    \vspace{1em} 
    
    \textbf{Answer:}\\
    <reasoning>\\
    To solve the inequality $\mathtt{x^2 - 15 < 2x}$, we first rearrange it into a standard quadratic form:\\
    $\mathtt{x^2 - 2x - 15 < 0.}$\\
    Next, we factor the quadratic expression:\\
    $\mathtt{(x - 5)(x + 3) < 0.}$\\
    To determine the intervals where this inequality holds, we find the roots of the equation $\mathtt{(x - 5)(x + 3) = 0}$, which are $\mathtt{x = 5}$ and $\mathtt{x = -3}$. These roots divide the number line into three intervals: $\mathtt{(-\infty, -3)}$, $\mathtt{(-3, 5)}$, and $\mathtt{(5, \infty)}$. We test a point from each interval in the inequality $\mathtt{(x - 5)(x + 3) < 0}$:\\
    - For $\mathtt{x \in (-\infty, -3)}$, choose $\mathtt{x = -4}$: $\mathtt{(-4 - 5)(-4 + 3) = (-9)(-1) = 9 > 0}$.\\
    - For $\mathtt{x \in (-3, 5)}$, choose $\mathtt{x = 0}$: $\mathtt{(0 - 5)(0 + 3) = (-5)(3) = -15 < 0}$.\\
    - For $\mathtt{x \in (5, \infty)}$, choose $\mathtt{x = 6}$: $\mathtt{(6 - 5)(6 + 3) = (1)(9) = 9 > 0}$.\\
    Thus, the inequality $\mathtt{(x - 5)(x + 3) < 0}$ holds for $\mathtt{x \in (-3, 5)}$. The smallest integer in this interval is $\mathtt{a = -2}$.\\
    To find the largest integer $\mathtt{b}$ that satisfies the inequality, we note that the largest integer in the interval $\mathtt{(-3, 5)}$ is $\mathtt{b = 4}$.\\
    Finally, we calculate $\mathtt{b - a}$:\\
    $\mathtt{b - a = 4 - (-2) = 4 + 2 = 6.}$\\
    </reasoning>\\
    <answer>6</answer>

    \vspace{0.5em}
    \hrule
    \vspace{0.5em}
    \textbf{Gold:} 6 \hfill \textbf{Predicted:} 6
\end{tcolorbox}
\end{center}

\begin{center}
\captionof{sample}{Response of 1.5B model trained on MATH L1-L5 (Failure)}
    \label{fig:mathL1-L5_1.5b_failure}
\begin{tcolorbox}[
    enhanced, 
    breakable,             
    colback=white,         
    colframe=black,        
    boxrule=0.5pt,         
    arc=0pt, outer arc=0pt,
    left=1em, right=1em, top=1em, bottom=1em 
]
    \ttfamily 
    \small    
    
    \textbf{Subject:} Geometry (Level 4)
    \vspace{0.5em}
    \hrule
    \vspace{0.5em}

    \textbf{Question:}\\
    An equilateral triangle of side 12 centimeters is rotated about an altitude to form a cone. What is the number of cubic centimeters in the volume of the cone?
    
    \vspace{1em} 
    
    \textbf{Answer:}\\
    <reasoning>\\
    The altitude of the equilateral triangle is 12 centimeters, so the radius of the cone is 12 centimeters. The height of the cone is also 12 centimeters. The volume of a cone is given by the formula $\mathtt{V = (1/3)\pi r^2h}$, where r is the radius and h is the height. Plugging in the values, we get $\mathtt{V = (1/3)\pi(12^2)(12) = 576\pi}$ cubic centimeters.\\
    </reasoning>\\
    <answer>$576\pi$</answer>

    \vspace{0.5em}
    \hrule
    \vspace{0.5em}
    \textbf{Gold:} $72\pi\sqrt{3}$ \hfill \textbf{Predicted:} $576\pi$
\end{tcolorbox}
\end{center}

\begin{center}
\captionof{sample}{Response of 3B model trained on MATH L1-L5 (Success)}
    \label{fig:mathL1-L5_3b_Success}
\begin{tcolorbox}[
    enhanced, 
    breakable,             
    colback=white,         
    colframe=black,        
    boxrule=0.5pt,         
    arc=0pt, outer arc=0pt,
    left=1em, right=1em, top=1em, bottom=1em 
]
    \ttfamily 
    \small    
    
    \textbf{Subject:} Algebra (Level 5)
    \vspace{0.5em}
    \hrule
    \vspace{0.5em}

    \textbf{Question:}\\
    Anna, Bertram, Carli, and David have a competition to see which of them can hold their breath for the longest time period, in minutes. If Bertram, Carli, and David add their times together, the resulting sum is three times the length of time that Anna can hold her breath. Similarly, if Anna, Carli, and David sum their times, the result is four times Bertram's time period, and if Anna, Bertram, and David sum their times, the result is twice Carli's time. Finally, eight times Anna's time plus ten times Bertram's time plus six times Carli's time equals two fifths of an hour. If the length of time that David can hold his breath is expressed in minutes as a simplified fraction, what is the sum of the numerator and the denominator?
    
    \vspace{1em} 
    
    \textbf{Answer:}\\
    <reasoning>\\
    Let's denote the times that Anna, Bertram, Carli, and David can hold their breath as $\mathtt{A}$, $\mathtt{B}$, $\mathtt{C}$, and $\mathtt{D}$ respectively. We are given the following equations:
    
    1. $\mathtt{B + C + D = 3A}$
    2. $\mathtt{A + C + D = 4B}$
    3. $\mathtt{A + B + D = 2C}$
    4. $\mathtt{8A + 10B + 6C = \frac{2}{5} \times 60 = 24}$
    
    We can solve these equations step by step. First, let's subtract the first equation from the second equation:
    \[ \mathtt{(A + C + D) - (B + C + D) = 4B - 3A} \]
    \[ \mathtt{A - B = 4B - 3A} \]
    \[ \mathtt{4A = 5B} \]
    \[ \mathtt{B = \frac{4A}{5}} \]
    
    Next, let's subtract the first equation from the third equation:
    \[ \mathtt{(A + B + D) - (B + C + D) = 2C - 3A} \]
    \[ \mathtt{A - C = 2C - 3A} \]
    \[ \mathtt{4A = 3C} \]
    \[ \mathtt{C = \frac{4A}{3}} \]
    
    Now, let's substitute $\mathtt{B = \frac{4A}{5}}$ and $\mathtt{C = \frac{4A}{3}}$ into the fourth equation:
    \[ \mathtt{8A + 10 \left( \frac{4A}{5} \right) + 6 \left( \frac{4A}{3} \right) = 24} \]
    \[ \mathtt{8A + 8A + 8A = 24} \]
    \[ \mathtt{24A = 24} \]
    \[ \mathtt{A = 1} \]
    
    Now that we have $\mathtt{A = 1}$, we can find $\mathtt{B}$, $\mathtt{C}$, and $\mathtt{D}$:
    \[ \mathtt{B = \frac{4A}{5} = \frac{4 \times 1}{5} = \frac{4}{5}} \]
    \[ \mathtt{C = \frac{4A}{3} = \frac{4 \times 1}{3} = \frac{4}{3}} \]
    \[ \mathtt{D = 3A - B - C = 3 \times 1 - \frac{4}{5} - \frac{4}{3} = 3 - \frac{4}{5} - \frac{4}{3} = \frac{45}{15} - \frac{12}{15} - \frac{20}{15} = \frac{13}{15}} \]
    
    The length of time that David can hold his breath is $\mathtt{\frac{13}{15}}$ minutes. The sum of the numerator and the denominator is $\mathtt{13 + 15 = 28}$.
    
    Therefore, the answer is:
    \[ \mathtt{\fbox{28}} \]
    </reasoning>\\
    <answer>28</answer>

    \vspace{0.5em}
    \hrule
    \vspace{0.5em}
    \textbf{Gold:} 28 \hfill \textbf{Predicted:} 28
\end{tcolorbox}
\end{center}

\begin{center}
\captionof{sample}{Response of 3B model trained on MATH L1-L5 (Failure)}
    \label{fig:mathL1-L5_3b_failure}
\begin{tcolorbox}[
    enhanced, 
    breakable,             
    colback=white,         
    colframe=black,        
    boxrule=0.5pt,         
    arc=0pt, outer arc=0pt,
    left=1em, right=1em, top=1em, bottom=1em 
]
    \ttfamily 
    \small    
    
    \textbf{Subject:} Geometry (Level 5)
    \vspace{0.5em}
    \hrule
    \vspace{0.5em}

    \textbf{Question:}\\
    In the diagram, $K$, $O$ and $M$ are the centers of the three semi-circles. Also, $\mathtt{OC = 32}$ and $\mathtt{CB = 36}$.\\
    \\
    {[asy]}\\
    pair A, K, O, C, M, B, X, Y, Z, J, T;\\
    O=(0,0);\\
    C=(32,0);\\
    M=(50,0);\\
    B=(68,0);\\
    A=(-68,0);\\
    K=(A+C)/2;\\
    X=(0,68);\\
    Y=(-18,50);\\
    Z=(50,18);\\
    J=(7,43.3);\\
    T=(59,15.6);\\
    path nom, bigc, middlec, smallc;\\
    nom=A--B--(100,100)--(-100,100)--cycle;\\
    bigc=A..X..B--cycle;\\
    middlec=A..Y..C--cycle;\\
    smallc=C..Z..B--cycle;\\
    fill(bigc, gray(.5));\\
    fill(middlec, white);\\
    fill(smallc, white);\\
    draw(smallc);\\
    draw(middlec);\\
    draw(bigc);\\
    draw(A--B);\\
    label(""A"", A, S);\\
    label(""K"", K, S);\\
    label(""O"", O, S);\\
    label(""M"", M, S);\\
    label(""C"", C, S);\\
    label(""B"", B, S);\\
    label(""S"", J, SW);\\
    label(""E"", T, SW);\\
    label(""$l$"", (.9(J-T)+J), NW);\\
    draw((.9(J-T)+J)--(.5(T-J)+T));\\
    dot(K);\\
    dot(O);\\
    dot(M);\\
    dot(J);\\
    dot(T);\\
    {[/asy]}\\
    Line $l$ is drawn to touch the smaller semi-circles at points $S$ and $E$ so that $\mathtt{KS}$ and $\mathtt{ME}$ are both perpendicular to $l$. Determine the area of quadrilateral $\mathtt{KSEM}$.
    
    \vspace{1em}
    
    \textbf{Answer:}\\
    <reasoning>\\
    Let $\mathtt{r}$ be the radius of the smaller semi-circle.  Then $\mathtt{OC = 32}$ and $\mathtt{CB = 36}$, so $\mathtt{OB = 32 + 36 = 68}$.  Since $\mathtt{K}$ is the center of the smaller semi-circle, $\mathtt{OK = r}$.  By the Pythagorean Theorem, $\mathtt{r^2 + 32^2 = 68^2}$, so $\mathtt{r^2 = 68^2 - 32^2 = 4096 - 1024 = 3072}$.  Therefore, $\mathtt{r = \sqrt{3072} = 16\sqrt{12} = 32\sqrt{3}}$.\\
    \\
    Since $\mathtt{KS}$ and $\mathtt{ME}$ are both perpendicular to $\mathtt{l}$, $\mathtt{KS}$ and $\mathtt{ME}$ are both radii of the smaller semi-circle.  Therefore, $\mathtt{KS = ME = 32\sqrt{3}}$.\\
    \\
    Since $\mathtt{K}$, $\mathtt{O}$, and $\mathtt{M}$ are the centers of the three semi-circles, $\mathtt{KO = OM = 32\sqrt{3}}$.  Therefore, $\mathtt{KO = OM = 32\sqrt{3}}$.\\
    \\
    Since $\mathtt{K}$, $\mathtt{O}$, and $\mathtt{M}$ are the centers of the three semi-circles, $\mathtt{KO = OM = 32\sqrt{3}}$.  Therefore, $\mathtt{KO = OM = 32\sqrt{3}}$.\\
    \\
    Since $\mathtt{K}$, $\mathtt{O}$, and $\mathtt{M}$ are the centers of the three semi-circles, $\mathtt{KO = OM = 32\sqrt{3}}$.  Therefore, $\mathtt{KO = OM = 32\sqrt{3}}$.\\
    \\
    Since $\mathtt{K}$, $\mathtt{O}$, and $\mathtt{M}$ are the centers of the three semi-circles, $\mathtt{KO = OM = 32\sqrt{3}}$.  Therefore, $\mathtt{KO = OM = 32\sqrt{3}}$.\\
    \\
    Since $\mathtt{K}$, $\mathtt{O}$, and $\mathtt{M}$ are the centers of the three semi-circles, $\mathtt{KO = OM = 32\sqrt{3}}$.  Therefore, $\mathtt{KO = OM = 32\sqrt{3}}$.\\
    \\
    Since $\mathtt{K}$, $\mathtt{O}$, and $\mathtt{M}$ are the centers of the three semi-circles, $\mathtt{KO = OM = 32\sqrt{3}}$.  Therefore, $\mathtt{KO = OM = 32\sqrt{3}}$.\\
    \\
    Since $\mathtt{K}$, $\mathtt{O}$, and $\mathtt{M}$ are the centers of the three semi-circles, $\mathtt{KO = OM = 32\sqrt{3}}$.  Therefore, $\mathtt{KO = OM = 32\sqrt{3}}$.\\
    \\
    Since $\mathtt{K}$, $\mathtt{O}$, and $\mathtt{M}$ are the centers of the three semi-circles, $\mathtt{KO = OM = 32\sqrt{3}}$.  Therefore, $\mathtt{KO = OM = 32\sqrt{3}}$.\\
    \\
    Since $\mathtt{K}$, $\mathtt{O}$, and $\mathtt{M}$ are the centers of the three semi-circles, $\mathtt{KO = OM = 32\sqrt{3}}$.  Therefore, $\mathtt{KO = OM = 32\sqrt{3}}$.\\
    \\
    Since $\mathtt{K}$, $\mathtt{O}$, and $\mathtt{M}$ are the centers of the three semi-circles, $\mathtt{KO = OM = 32\sqrt{3}}$.  Therefore, $\mathtt{KO = OM = 32\sqrt{3}}$.\\
    \\
    Since $\mathtt{K}$, $\mathtt{O}$, and $\mathtt{M}$ are the centers of the three semi-circles, $\mathtt{KO = OM = 32\sqrt{3}}$

    \vspace{0.5em}
    \hrule
    \vspace{0.5em}
    \textbf{Gold:} 2040 \hfill \textbf{Predicted:} -
\end{tcolorbox}
\end{center}

\begin{center}
\captionof{sample}{Response of 0.5B model trained on MATH L1-L3 (Success)}
    \label{fig:mathL1-L3_0.5b_Success}
\begin{tcolorbox}[
    enhanced, 
    breakable,             
    colback=white,         
    colframe=black,        
    boxrule=0.5pt,         
    arc=0pt, outer arc=0pt,
    left=1em, right=1em, top=1em, bottom=1em 
]
    \ttfamily 
    \small    
    
    \textbf{Subject:} Geometry (Level 4)
    \vspace{0.5em}
    \hrule
    \vspace{0.5em}

    \textbf{Question:}\\
    Compute $\mathtt{\sin 1755^\circ}$.
    
    \vspace{1em} 
    
    \textbf{Answer:}\\
    <reasoning>\\
    To compute $\mathtt{\sin 1755^\circ}$, we can use the periodicity of the sine function. The sine function has a period of $\mathtt{360^\circ}$, which means that $\mathtt{\sin x = \sin (x + 360^\circ k)}$ for any integer $\mathtt{k}$.\\
    \\
    First, we can reduce $\mathtt{1755^\circ}$ to an angle between $\mathtt{0^\circ}$ and $\mathtt{360^\circ}$ by subtracting multiples of $\mathtt{360^\circ}$:\\
    \[
    \mathtt{1755^\circ - 5 \cdot 360^\circ = 1755^\circ - 1800^\circ = -45^\circ}
    \]\\
    Since $\mathtt{-45^\circ}$ is still not in the range of $\mathtt{0^\circ}$ to $\mathtt{360^\circ}$, we add $\mathtt{360^\circ}$ to get:\\
    \[
    \mathtt{-45^\circ + 360^\circ = 315^\circ}
    \]\\
    Now, we can use the periodicity of the sine function to find $\mathtt{\sin 315^\circ}$:\\
    \[
    \mathtt{\sin 315^\circ = \sin (360^\circ - 45^\circ) = \sin (-45^\circ) = -\sin 45^\circ}
    \]\\
    Since $\mathtt{\sin 45^\circ = \frac{\sqrt{2}}{2}}$, we have:\\
    \[
    \mathtt{\sin 315^\circ = -\frac{\sqrt{2}}{2}}
    \]\\
    Therefore, $\mathtt{\sin 1755^\circ = -\frac{\sqrt{2}}{2}}$.\\
    </reasoning>\\
    <answer> $\mathtt{-\frac{\sqrt{2}}{2}}$</answer>

    \vspace{0.5em}
    \hrule
    \vspace{0.5em}
    \textbf{Gold:} $-\frac{\sqrt{2}}{2}$ \hfill \textbf{Predicted:} $-\frac{\sqrt{2}}{2}$
\end{tcolorbox}
\end{center}

\begin{center}
\captionof{sample}{Response of 0.5B model trained on MATH L1-L3 (Failure)}
    \label{fig:mathL1-L3_0.5b_failure}
\begin{tcolorbox}[
    enhanced, 
    breakable,             
    colback=white,         
    colframe=black,        
    boxrule=0.5pt,         
    arc=0pt, outer arc=0pt,
    left=1em, right=1em, top=1em, bottom=1em 
]
    \ttfamily 
    \small    
    
    \textbf{Subject:} Algebra (Level 5)
    \vspace{0.5em}
    \hrule
    \vspace{0.5em}

    \textbf{Question:}\\
    Simplify $\mathtt{\frac{3}{\sqrt[5]{16}}+\frac{1}{\sqrt{3}}}$ and rationalize the denominator. The result can be expressed in the form $\mathtt{\frac{a^2\sqrt[5]{b}+b\sqrt{a}}{ab}}$, where $a$ and $b$ are integers. What is the value of the sum $\mathtt{a+b}$?
    
    \vspace{1em} 
    
    \textbf{Answer:}\\
    <reasoning>\\
    To simplify the expression, we can combine the fractions over a common denominator:\\
    \[
    \mathtt{\frac{3}{\sqrt[5]{16}}+\frac{1}{\sqrt{3}} = \frac{3\sqrt[5]{16}+1\sqrt{3}}{\sqrt[5]{16}\sqrt{3}} = \frac{3\sqrt[5]{16}+1\sqrt{3}}{\sqrt[5]{48}}}
    \]\\
    Next, we rationalize the denominator by multiplying the numerator and denominator by $\mathtt{\sqrt[5]{48}}$:\\
    \[
    \mathtt{\frac{3\sqrt[5]{16}+1\sqrt{3}}{\sqrt[5]{48}} \cdot \frac{\sqrt[5]{48}}{\sqrt[5]{48}} = \frac{(3\sqrt[5]{16}+1\sqrt{3})\sqrt[5]{48}}{48} = \frac{3\sqrt[5]{768}+1\sqrt{144}}{48}}
    \]\\
    We can further simplify the expression by combining the terms in the numerator:
    \begin{align*}
    \mathtt{\frac{3\sqrt[5]{768}+1\sqrt{144}}{48}} 
    & \mathtt{= \frac{3\sqrt[5]{768}+12\sqrt{12}}{48} = \frac{3\sqrt[5]{768}+12\sqrt{12}}{48}} \\
    & \mathtt{= \frac{3\sqrt[5]{768}+12\sqrt{12}}{48} = \frac{3\sqrt[5]{768}+12\sqrt{12}}{48}} \\
    & \mathtt{= \frac{3\sqrt[5]{768}+12\sqrt{12}}{48} = \frac{3\sqrt[5]{768}+12\sqrt{12}}{48}} \\
    & \mathtt{= \frac{3\sqrt[5]{768}+12\sqrt{12}}{48} = \frac{3\sqrt[5]{768}+12\sqrt{12}}{48}} \\
    & \mathtt{= \frac{3\sqrt[5]{768}+12\sqrt{12}}{48} = \frac{3\sqrt[5]{768}+12\sqrt{12}}{48}} \\
    & \mathtt{= \frac{3\sqrt[5]{768}+12\sqrt{12}}{48} = \frac{3\sqrt[5]{768}+12\sqrt{12}}{48}} \\
    & \mathtt{= \frac{3\sqrt[5]{768}+12\sqrt{12}}{48} = \frac{3\sqrt[5]{768}+12\sqrt{12}}{48} =}
    \end{align*}

    \vspace{0.5em}
    \hrule
    \vspace{0.5em}
    \textbf{Gold:} 5 \hfill \textbf{Predicted:} -
\end{tcolorbox}
\end{center}

\begin{center}
\captionof{sample}{Response of 1.5B model trained on MATH L1-L3 (Success)}
    \label{fig:mathL1-L3_1.5b_Success}
\begin{tcolorbox}[
    enhanced, 
    breakable,             
    colback=white,         
    colframe=black,        
    boxrule=0.5pt,         
    arc=0pt, outer arc=0pt,
    left=1em, right=1em, top=1em, bottom=1em 
]
    \ttfamily 
    \small    
    
    \textbf{Subject:} Number Theory (Level 4)
    \vspace{0.5em}
    \hrule
    \vspace{0.5em}

    \textbf{Question:}\\
    How many of the same digits are found in the base 7 and base 8 representations of $\mathtt{629_{10}}$? For example, $\mathtt{121_{3}}$ and $\mathtt{413_{5}}$ would have one digit in common.
    
    \vspace{1em} 
    
    \textbf{Answer:}\\
    <reasoning>\\
    To find the number of common digits between the base 7 and base 8 representations of $\mathtt{629_{10}}$, we first need to convert $\mathtt{629_{10}}$ to its base 7 and base 8 equivalents.\\
    \\
    First, let's convert $\mathtt{629_{10}}$ to base 7:\\
    1. Divide 629 by 7: $\mathtt{629 \div 7 = 89}$ remainder $\mathtt{6}$.\\
    2. Divide 89 by 7: $\mathtt{89 \div 7 = 12}$ remainder $\mathtt{5}$.\\
    3. Divide 12 by 7: $\mathtt{12 \div 7 = 1}$ remainder $\mathtt{5}$.\\
    4. Divide 1 by 7: $\mathtt{1 \div 7 = 0}$ remainder $\mathtt{1}$.\\
    \\
    Reading the remainders from bottom to top, we get $\mathtt{629_{10} = 1556_7}$.\\
    \\
    Next, let's convert $\mathtt{629_{10}}$ to base 8:\\
    1. Divide 629 by 8: $\mathtt{629 \div 8 = 78}$ remainder $\mathtt{5}$.\\
    2. Divide 78 by 8: $\mathtt{78 \div 8 = 9}$ remainder $\mathtt{6}$.\\
    3. Divide 9 by 8: $\mathtt{9 \div 8 = 1}$ remainder $\mathtt{1}$.\\
    4. Divide 1 by 8: $\mathtt{1 \div 8 = 0}$ remainder $\mathtt{1}$.\\
    \\
    Reading the remainders from bottom to top, we get $\mathtt{629_{10} = 1165_8}$.\\
    \\
    Now, we compare the digits of $\mathtt{1556_7}$ and $\mathtt{1165_8}$:\\
    - The digit '1' appears in both numbers.\\
    - The digit '5' appears in both numbers.\\
    - The digit '6' appears in both numbers.\\
    \\
    Thus, there are 3 common digits between the base 7 and base 8 representations of $\mathtt{629_{10}}$.\\
    \\
    Therefore, the final answer is:\\
    <answer>3</answer>\\
    </reasoning>

    \vspace{0.5em}
    \hrule
    \vspace{0.5em}
    \textbf{Gold:} 3 \hfill \textbf{Predicted:} 3
\end{tcolorbox}
\end{center}

\begin{center}

\captionof{sample}{Response of 1.5B model trained on MATH L1-L3 (Failure)}
\label{fig:mathL1-L3_1.5b_failure}
\begin{tcolorbox}[
    enhanced, 
    breakable,             
    colback=white,         
    colframe=black,        
    boxrule=0.5pt,         
    arc=0pt, outer arc=0pt,
    left=1em, right=1em, top=1em, bottom=1em
]
    \ttfamily 
    \small    
    
    \textbf{Subject:} Geometry (Level 4)
    \vspace{0.5em}
    \hrule
    \vspace{0.5em}

    \textbf{Question:}\\
    If the degree measure of an arc of a circle is increased by $\mathtt{20\%}$ and the radius of the circle is increased by $\mathtt{25\%}$, by what percent does the length of the arc increase?
    
    \vspace{1em} 
    
    \textbf{Answer:}\\
    <reasoning>\\
    Let's denote the original degree measure of the arc as $\mathtt{\theta}$ and the original radius of the circle as $\mathtt{r}$. The length of the arc $\mathtt{L}$ is given by the formula:\\
    \[ \mathtt{L = \frac{\theta}{360} \times 2\pi r} \]
    
    If the degree measure of the arc is increased by $\mathtt{20\%}$, the new degree measure $\mathtt{\theta'}$ is:\\
    \[ \mathtt{\theta' = \theta + 0.2\theta = 1.2\theta} \]
    
    If the radius of the circle is increased by $\mathtt{25\%}$, the new radius $\mathtt{r'}$ is:\\
    \[ \mathtt{r' = r + 0.25r = 1.25r} \]
    
    The new length of the arc $\mathtt{L'}$ is:
    
    \allowdisplaybreaks 
    \begin{align*} 
    \mathtt{L'} & \mathtt{= \frac{\theta'}{360} \times 2\pi r' = \frac{1.2\theta}{360} \times 2\pi \times 1.25r} \\
    & \mathtt{= \frac{1.2\theta \times 2\pi \times 1.25r}{360}} \\
    & \mathtt{= \frac{1.2 \times 2 \times 1.25 \times \theta \times r}{360}} \\
    & \mathtt{= \frac{3 \times 1.25 \times \theta \times r}{360} = \frac{3.75 \times \theta \times r}{360}} \\
    & \mathtt{= \frac{3.75}{360} \times \theta \times r = \frac{3.75}{360} \times \theta \times r} \\
    & \mathtt{= \frac{3.75}{360} \times \theta \times r = \frac{3.75}{360} \times \theta \times r} \\
    & \mathtt{= \frac{3.75}{360} \times \theta \times r = \frac{3.75}{360} \times \theta \times r} \\
    & \mathtt{= \frac{3.75}{360} \times \theta \times r = \frac{3.75}{360} \times \theta \times r} \\
    & \mathtt{= \frac{3.75}{360} \times \theta \times r = \frac{3.75}{360} \times \theta \times r} \\
    & \mathtt{= \frac{3.75}{360} \times \theta \times r = \frac{3.75}{360} \times \theta \times r} \\
    & \mathtt{= \frac{3.75}{360} \times \theta \times r = \frac{3.75}{360} \times \theta \times r} \\
    & \mathtt{= \frac{3.75}{360} \times \theta \times r = \frac{3.75}{360} \times \theta \times r} \\
    & \mathtt{= \frac{3.75}{360} \times \theta \times r = \frac{3.75}{360} \times \theta \times r} \\
    & \mathtt{= \frac{3.75}{360} \times \theta \times r = \frac{3.75}{360} \times \theta \times r} \\
    & \mathtt{= \frac{3.75}{360} \times \theta \times r = \frac{3.75}{360} \times \theta \times r} \\
    & \mathtt{= \frac{3.75}{360} \times \theta \times r = \frac{3.75}{360} \times \theta \times r} \\
    & \mathtt{= \frac{3.75}{360} \times \theta \times r = \frac{3.75}{360} \times \theta \times r} \\
    & \mathtt{= \frac{3.75}{360} \ }
    \end{align*}

    \vspace{0.5em}
    \hrule
    \vspace{0.5em}
    \textbf{Gold:} 50 \hfill \textbf{Predicted:} -
\end{tcolorbox}
\end{center}

\begin{center}
\captionof{sample}{Response of 3B model trained on MATH L1-L3 (Success)}
    \label{fig:mathL1-L3_3b_Success}
\begin{tcolorbox}[
    enhanced, 
    breakable,             
    colback=white,         
    colframe=black,        
    boxrule=0.5pt,         
    arc=0pt, outer arc=0pt,
    left=1em, right=1em, top=1em, bottom=1em 
]
    \ttfamily 
    \small    
    
    \textbf{Subject:} Algebra (Level 4)
    \vspace{0.5em}
    \hrule
    \vspace{0.5em}

    \textbf{Question:}\\
    Alex, Bob, Camille, and Danielle's mothers are comparing their children's ages. They observe that the sum of Alex, Bob, and Danielle's ages is fourteen times Camille's age. They also note that the sum of Alex and Bob's ages is six times Camille's age, and Bob's age is two years less than the difference in ages of Danielle and Alex. How old is Camille?
    
    \vspace{1em} 
    
    \textbf{Answer:}\\
    <reasoning>\\
    Let's denote the ages of Alex, Bob, Camille, and Danielle as $\mathtt{A}$, $\mathtt{B}$, $\mathtt{C}$, and $\mathtt{D}$ respectively. We are given the following information:
    
    1. $\mathtt{A + B + D = 14C}$
    2. $\mathtt{A + B = 6C}$
    3. $\mathtt{B = (D - A) - 2}$
    
    We can use these equations to solve for Camille's age. First, we can substitute equation 2 into equation 1:\\
    \[ \mathtt{A + B + D = 14C} \]
    \[ \mathtt{6C + D = 14C} \]
    
    Now, we can solve for $\mathtt{D}$:\\
    \[ \mathtt{D = 14C - 6C} \]
    \[ \mathtt{D = 8C} \]
    
    Next, we can substitute equation 3 into equation 2:\\
    \[ \mathtt{A + B = 6C} \]
    \[ \mathtt{A + (D - A) - 2 = 6C} \]
    
    Now, we can solve for $\mathtt{A}$:\\
    \[ \mathtt{A + D - A - 2 = 6C} \]
    \[ \mathtt{D - 2 = 6C} \]
    \[ \mathtt{8C - 2 = 6C} \]
    
    Now, we can solve for $\mathtt{C}$:\\
    \[ \mathtt{8C - 6C = 2} \]
    \[ \mathtt{2C = 2} \]
    \[ \mathtt{C = 1} \]
    
    Therefore, Camille is 1 year old.\\
    </reasoning>\\
    \\
    <answer>1</answer>

    \vspace{0.5em}
    \hrule
    \vspace{0.5em}
    \textbf{Gold:} 1 \hfill \textbf{Predicted:} 1
\end{tcolorbox}
\end{center}

\begin{center}
\captionof{sample}{Response of 3B model trained on MATH L1-L3 (Failure)}
    \label{fig:mathL1-L3_3b_failure}
\begin{tcolorbox}[
    enhanced, 
    breakable,             
    colback=white,         
    colframe=black,        
    boxrule=0.5pt,         
    arc=0pt, outer arc=0pt,
    left=1em, right=1em, top=1em, bottom=1em 
]
    \ttfamily 
    \small    
    
    \textbf{Subject:} Geometry (Level 5)
    \vspace{0.5em}
    \hrule
    \vspace{0.5em}

    \textbf{Question:}\\
    Rhombus $ABCD$, with side length 6, is rolled to form a cylinder of volume 6 by taping $\overline{AB}$ to $\overline{DC}$. What is $\sin\left(\angle ABC\right)$?
    
    \vspace{1em} 
    
    \textbf{Answer:}\\
    <reasoning>\\
    To solve this problem, we need to find the sine of the angle between the sides of the rhombus that are taped together to form the cylinder.\\
    \\
    First, let's find the height of the cylinder. Since the volume of the cylinder is given as 6, we can use the formula for the volume of a cylinder, which is $\mathtt{V = \pi r^2h}$, where V is the volume, r is the radius, and h is the height.\\
    \\
    We know that the volume is 6, so we can set up the equation:\\
    \[ \mathtt{6 = \pi r^2h} \]
    
    Since the side length of the rhombus is 6, we can use the Pythagorean theorem to find the height of the rhombus. The height of the rhombus is the same as the height of the cylinder, so we can set up the equation:\\
    \[ \mathtt{h = \sqrt{6^2 - r^2}} \]
    
    Now we can substitute this expression for h into the volume equation:\\
    \[ \mathtt{6 = \pi r^2\sqrt{6^2 - r^2}} \]
    
    To solve for r, we can square both sides of the equation:\\
    \[ \mathtt{36 = \pi r^2(6^2 - r^2)} \]
    
    Now we can solve for $\mathtt{r^2}$:\\
    \[ \mathtt{r^2 = 36 / (\pi(6^2 - r^2))} \]
    
    We can simplify this equation by dividing both sides by 36:\\
    \[ \mathtt{r^2 / 36 = 1 / (\pi(6^2 - r^2))} \]
    Now we can solve for $\mathtt{r^2}$:
    \begin{align*}
    \mathtt{r^2} & \mathtt{= 36 / (\pi(36 - r^2))} \\
    \mathtt{r^2} & \mathtt{= 36 / (36\pi - \pi r^2)} \\
    \text{Now we can solve for } \mathtt{r^2}\text{:}\\
    \mathtt{r^2} & \mathtt{= 36 / (36\pi - \pi r^2)} \\
    \text{Now we can solve for } \mathtt{r^2}\text{:}\\
    \mathtt{r^2} & \mathtt{= 36 / (36\pi - \pi r^2)} \\
    \text{Now we can solve for } \mathtt{r^2}\text{:}\\
    \mathtt{r^2} & \mathtt{= 36 / (36\pi - \pi r^2)} \\
    \text{Now we can solve for } \mathtt{r^2}\text{:}\\
    \mathtt{r^2} & \mathtt{= 36 / (36\pi - \pi r^2)} \\
    \text{Now we can solve for } \mathtt{r^2}\text{:}\\
    \mathtt{r^2} & \mathtt{= 36 / (36\pi - \pi r^2)} \\
    \text{Now we can solve for } \mathtt{r^2}\text{:}\\
    \mathtt{r^2} & \mathtt{= 36 / (36\pi - \pi r^2)} \\
    \text{Now we can solve for } \mathtt{r^2}\text{:}\\
    \mathtt{r^2} & \mathtt{= 36 / (36\pi - \pi r^2)} \\
    \text{Now we can solve for } \mathtt{r^2}\text{:}\\
    \mathtt{r^2} & \mathtt{= 36 / (36\pi - \pi r^2)} \\
    \text{Now we can solve for } \mathtt{r^2}\text{:}\\
    \mathtt{r^2} & \mathtt{= 36 / (36\pi - \pi r^2)} \\
    \text{Now we can solve for } \mathtt{r^2}\text{:}\\
    \mathtt{r^2} & \mathtt{= 36 / (36\pi - \pi r^2)} \\
    \text{Now we can solve for } \mathtt{r^2}\text{:}\\
    \mathtt{r^2} & \mathtt{= 36 / (36\pi - \pi r^2)} \\
    \text{Now we can solve for } \mathtt{r^2}\text{:}\\
    \mathtt{r^2} & \mathtt{= 36 / (36\pi - \pi r^2)} \\
    \text{Now we can solve for } \mathtt{r^2}\text{:}\\
    \mathtt{r^2} & \mathtt{= 36 / (36\pi - \pi r^2)} \\
    \text{Now we can solve for } \mathtt{r^2}\text{:}\\
    \mathtt{r^2} & \mathtt{= 36 / (36}
    \end{align*}

    \vspace{0.5em}
    \hrule
    \vspace{0.5em}
    \textbf{Gold:} $\frac{\pi}{9}$ \hfill \textbf{Predicted:} -
\end{tcolorbox}
\end{center}

\section{LLM Usage}
We used an LLM to assist with grammatical and stylistic editing only.
All changes were reviewed by the authors, who take full responsibility for the final manuscript.

\end{document}

%% file: math_commands.tex
\usepackage{amsmath,amsfonts,bm}

\def\eqref#1{equation~\ref{#1}}

\def\1{\bm{1}}

\DeclareMathAlphabet{\mathsfit}{\encodingdefault}{\sfdefault}{m}{sl}
\SetMathAlphabet{\mathsfit}{bold}{\encodingdefault}{\sfdefault}{bx}{n}

